\documentclass[11pt]{article}

\usepackage{fontspec}

\usepackage[final]{acl}

\usepackage{booktabs}
\usepackage{multirow}
\usepackage{tabularx}
\usepackage{array}
\usepackage{colortbl}
\usepackage{float}

\usepackage{graphicx}

\usepackage{amsmath}
\usepackage{amssymb}

\usepackage{pifont}

\usepackage{microtype}
\usepackage[normalem]{ulem}
\usepackage{ragged2e}

\usepackage[title]{appendix}

\usepackage[english,bidi=default]{babel}

\babelfont{rm}[
    BoldFont=TeXGyreTermesX-Bold.otf,
    ItalicFont=TeXGyreTermesX-Italic.otf,
    BoldItalicFont=TeXGyreTermesX-BoldItalic.otf
]{TeXGyreTermesX-Regular.otf}

\babelprovide[import]{hindi}
\babelfont[hindi]{rm}[
    Path=fonts/,
    BoldFont=NotoSerifDevanagari-Bold.ttf
]{NotoSerifDevanagari-Regular.ttf}

\babelprovide[import,onchar=ids fonts]{telugu}
\babelfont[telugu]{rm}[
    Path=fonts/,
    BoldFont=NotoSerifTelugu-Bold.ttf
]{NotoSerifTelugu-Regular.ttf}

\babelprovide[import]{tamil}
\babelfont[tamil]{rm}[
    Path=fonts/,
    BoldFont=NotoSerifTamil-Bold.ttf
]{NotoSerifTamil-Regular.ttf}
\newcommand{\TamilText}[1]{\foreignlanguage{tamil}{#1}}

\newcommand{\ParaMark}{\texttt{\textbackslash n\textbackslash n}}

\newcommand{\cmark}{\ding{51}}
\newcommand{\xmark}{\ding{55}}

\definecolor{avgHigh}{HTML}{C6EFCE}   
\definecolor{avgMid}{HTML}{FFEB9C}    
\definecolor{avgLow}{HTML}{FFC7CE}    

\newcommand{\hi}[1]{\cellcolor{avgHigh}#1}
\newcommand{\md}[1]{\cellcolor{avgMid}#1}
\newcommand{\lo}[1]{\cellcolor{avgLow}#1}

\newcommand{\best}[1]{\textbf{\uline{#1}}}
\newcommand{\secbest}[1]{\textbf{\textit{#1}}}

\title{IndicDetect: Evaluating Cross-Lingual LLM-Generated Text Detection for Hindi, Telugu, and Tamil}

\author{Bhaskar Ganesh Devalla$^{1}$\thanks{~Equal contribution} ~~~ Junchao Wu$^{1}$\footnotemark[1] ~~~ Nilesh Dokuparthi$^{2}$ ~~~ Greeshma Yaluru$^{3}$\\
\bf Tatiana Muniz Rodriguez$^{1}$ ~~~ \bf Lidia S. Chao$^{1}$ ~~~ \bf Derek F. Wong$^{1}$\thanks{Corresponding author.}\\
$^{1}$NLP$^2$CT Lab, Faculty of Information Science and Computing, University of Macau\\
$^{2}$Department of Applied Computer Science, SRH University, Germany\\
$^{3}$Populus Group, Pittsburgh, United States of America\\
\texttt{nlp2ct.\{bhaskarganeshdevalla,junchao,tatiana\}@gmail.com}\\
\texttt{\{derekfw,lidiasc\}@um.edu.mo} ~~~ \texttt{\{gyaluru,dnileshmohan\}@gmail.com} 
}

\begin{document}
\maketitle

\begin{abstract}
The rapid proliferation of LLMs has further heightened the need to develop
dependable AI-generated text detection, especially beyond English. Nevertheless,
current benchmarks pay little attention to Indic languages and test detectors in
idealized settings that do not represent the real world. We present a generalized
benchmark for AI-generated text detection in Hindi, Telugu, and Tamil, which we call
\textbf{\textit{IndicDetect}}, designed to assess the robustness of detectors under
realistic distribution shifts. \textbf{\textit{IndicDetect}} comprises highly curated
human-written texts matched with LLM-generated counterparts across various domains
and generators, and systematically evaluates detectors in the presence of domain
shift, generator shift, and adversarial perturbation. Using a single and repeatable
evaluation scheme, we evaluate a wide range of statistical and neural detectors.
We find substantial robustness failures: supervised neural detectors perform well
in-distribution, while training-free methods degrade considerably under unseen
generators and adversarial attacks. The severity of these failures varies across
languages, with Hindi exhibiting the largest overall degradation under adversarial
perturbations. These results highlight that the primary weakness of existing
detectors in Indic settings lies in their robustness, not in their peak accuracy.
\textbf{\textit{IndicDetect}} provides standard data splits, an evaluation protocol,
and baselines to establish a robust, language-aware foundation for AI-generated
text detection in Indic scripts.\footnote{Code and Data:
\url{https://github.com/NLP2CT/IndicDetect}}
\end{abstract}

\section{Introduction}
Large language models (LLMs) have gained widespread popularity for producing
coherent and fluent text in various applications, raising concerns about
plagiarism, misinformation, and misuse
\citep{ippolito-etal-2020-automatic}.
This has led to the development of reliable AI-generated text detection as an
active research area~\citep{wu-etal-2025-survey}.
Although current research has sought to advance this field, available benchmarks
and evaluations are predominantly English-centric and are often constructed under
idealized conditions that do not reflect real-world application
contexts~\citep{10.1007/978-981-95-3352-7_21}. These limitations are especially
severe for Indic languages such as Hindi, Telugu, and Tamil
\citep{goyal-etal-2022-flores}, which possess distinct linguistic characteristics
and are underrepresented in existing detection standards. Furthermore, detectors
that perform well under matched conditions repeatedly fail under domain shifts,
unseen generators, or adversarial perturbations, casting doubt on their robustness
and real-world reliability. To address these gaps, we propose
\textbf{\textit{IndicDetect}}, a benchmark for AI-generated text detection in Hindi,
Telugu, and Tamil, designed to evaluate detector robustness under realistic
distribution shifts. \textbf{\textit{IndicDetect}} pairs carefully curated
human-written texts with LLM-generated counterparts across multiple domains and
generators, and evaluates a range of statistical and neural detectors under a
single reproducible protocol. Our findings show that robustness, rather than the
highest accuracy in-distribution, is the primary limitation of current detectors
in Indic languages. Robustness failures vary substantially across languages and
evaluation conditions, with Hindi showing the greatest overall vulnerability to
adversarial perturbations.\looseness=-1

\section{Related Work}\label{sec:setup}

\begin{table*}[t]
\small
\centering
\setlength{\tabcolsep}{4.2pt} 
\renewcommand{\arraystretch}{1.08}
\begin{tabular}{@{}l|c|c|c|c|c|c@{}}
\toprule
\textbf{Name} & \textbf{Size} & \textbf{Langs} & \textbf{Multi-Domain} &
\textbf{Multi-Generator} & \textbf{Adversarial} & \textbf{Indic} \\
\midrule
TuringBench \citep{uchendu2021turingbenchbenchmarkenvironmentturing} & 200k  & EN    & \xmark & \cmark & \xmark & \xmark \\
RuATD \citep{Shamardina2022RuATD}                                   & 215k  & RU    & \cmark & \cmark & \xmark & \xmark \\
HC3 \citep{guo-etal-2023-hc3}                                       & 26.9k & EN/ZH & \cmark & \xmark & \xmark & \xmark \\
MGTBench \citep{he2024mgtbenchbenchmarkingmachinegeneratedtext}     & 2.8k  & EN    & \cmark & \cmark & \cmark & \xmark \\
MULTITuDE \citep{Macko_2023}                                        & 74.1k & Multi & \cmark & \cmark & \xmark & \xmark \\
AuText2023 \citep{sarvazyan2023autextification}                     & 160k  & EN/ES & \cmark & \xmark & \xmark & \xmark \\
M4 \citep{wang2024m4multigeneratormultidomainmultilingual}          & 122k  & Multi & \cmark & \cmark & \xmark & \xmark \\
CCD \citep{Wang2023CCD}                                             & 467k  & EN    & \xmark & \cmark & \cmark & \xmark \\
IMDGSP \citep{mosca-etal-2023-distinguishing}                       & 29k   & EN    & \xmark & \cmark & \xmark & \xmark \\
HC-Var \citep{xu2023generalization}                                 & 145k  & EN    & \cmark & \xmark & \xmark & \xmark \\
HC3 Plus \citep{Su2023HC3Plus}                                      & 210k  & EN/ZH & \cmark & \xmark & \xmark & \xmark \\
MAGE \citep{li2023magemaskedgenerativeencoder}                      & 447k  & EN    & \cmark & \cmark & \xmark & \xmark \\
HC3-French \citep{antoun-etal-2023-towards}                         & 54k   & FR    & \xmark & \xmark & \cmark & \xmark \\
DetectRL \citep{wu2025detectrlbenchmarkingllmgeneratedtext}         & 1.4M  & EN    & \cmark & \cmark & \cmark & \xmark \\
RAID \citep{dugan-etal-2024-raid}                                   & 6.2M  & EN    & \cmark & \cmark & \cmark & \xmark \\
DetectRL-X \citep{wu-etal-2026-detectrl}                                 & 3.46M & Multi & \cmark & \cmark & \cmark & \xmark \\
\midrule
\textbf{IndicDetect (Ours)} & \textbf{84k} & \textbf{HI/TE/TA} &
\textbf{\cmark} & \textbf{\cmark} & \textbf{\cmark} & \textbf{\cmark} \\
\bottomrule
\end{tabular}
\caption{Comparison of AI-generated text detection benchmarks across domain
generalization, generator generalization, adversarial robustness, and language
coverage. \textbf{\textit{IndicDetect}} is the only benchmark targeting Indic scripts
while supporting all evaluation axes.}
\label{tab:indicdetect-benchmarks}
\end{table*}
LLMs have transformed natural language processing by generating fluent,
coherent, and contextually relevant text across diverse domains. However,
this capability introduces significant risks: LLM-generated text can be
exploited for plagiarism, the spread of factual misinformation in journalism,
or the automation of propaganda on social media \citep{wu2025detectrlbenchmarkingllmgeneratedtext}.
The increasing difficulty of manually distinguishing LLM-generated from
human-written text makes automated detection a pressing research priority, particularly as LLM-generated content increasingly permeates 
low-resource and morphologically rich languages. To address these threats, researchers have developed supervised classifiers
and zero-shot statistical detectors~\citep{mitchell2023detectgptzeroshotmachinegeneratedtext,DBLP:conf/coling/WuZWY0CZ25,DBLP:journals/tacl/ChenWYZWLWYCW25}. While these
methods perform well under controlled conditions, their effectiveness
degrades substantially when applied to noisy or domain-shifted data.
DetectGPT, for instance, exploits probability curvature to identify
machine-generated text, but fails when the text is paraphrased or
adversarially perturbed, revealing a fundamental gap between research prototypes
and practical deployment. To measure progress systematically, several benchmarks have been established,
including TuringBench \citep{uchendu2021turingbenchbenchmarkenvironmentturing},
MGTBench \citep{he2024mgtbenchbenchmarkingmachinegeneratedtext}, MULTITuDE
\citep{Macko_2023}, MAGE \citep{li2023magemaskedgenerativeencoder}, and M4
\citep{wang2024m4multigeneratormultidomainmultilingual}. These benchmarks
compare detectors across domains, LLMs, and languages and provide useful
reference points. Nevertheless, they are largely idealized in their
construction: machine-generated and human-written texts are kept in clean,
separate pools, without accounting for human editing, noise, or adversarial
interference. As a result, detectors that succeed on these benchmarks often
fail to generalize to real-world settings. A comparative summary of existing
benchmarks and their limitations is provided in
Table~\ref{tab:indicdetect-benchmarks}. RAID \citep{dugan-etal-2024-raid} pushes detector robustness evaluation
further with a large-scale benchmark of over six million LLM-generated texts
spanning multiple models, topics, and decoding strategies, revealing that
most existing detectors operate reliably only under narrow conditions and
break under even minor surface edits. DetectRL
\citep{wu2025detectrlbenchmarkingllmgeneratedtext} provides a realistic
evaluation of AI-generated text detection across multiple domains, generators,
text lengths, and adversarial conditions, demonstrating that most zero-shot
detectors fail substantially under paraphrasing, perturbations, and domain or
model shifts, while supervised classifiers achieve stronger results.
A recent multilingual extension, DetectRL-X \citep{wu-etal-2026-detectrl}, broadens coverage to eight languages spanning English, Chinese, Spanish, Arabic, French, Russian, Portuguese, and German, yet still covers no Brahmic-script language, leaving Hindi, Telugu, and Tamil unaddressed. Concurrent efforts such as the Counter Turing Test
\citep{chakraborty-etal-2023-counter} and Text Guardian
\citep{10.1007/978-981-96-4151-2_34} target Hindi news detection in a
monolingual setting. \textbf{\textit{IndicDetect}} extends this line of work to three
Indic languages, four domains, and Brahmic-script adversarial perturbations
(Table~\ref{tab:indicdetect-benchmarks}). Unlike DetectRL
\citep{wu2025detectrlbenchmarkingllmgeneratedtext}, \textbf{\textit{IndicDetect}}
introduces script-specific perturbations targeting Brahmic tokenization
patterns, evaluates multiple Indic languages under identical settings to
isolate cross-language robustness gaps, and enforces monolingual,
script-constant evaluation. These distinctions make \textbf{\textit{IndicDetect}} a uniquely positioned resource
for advancing robust AI-generated text detection in morphologically rich
Indic languages and related scripts. \looseness=-1

\begin{figure*}[t]
    \centering
    \vspace{-6pt}
    \includegraphics[width=\textwidth]{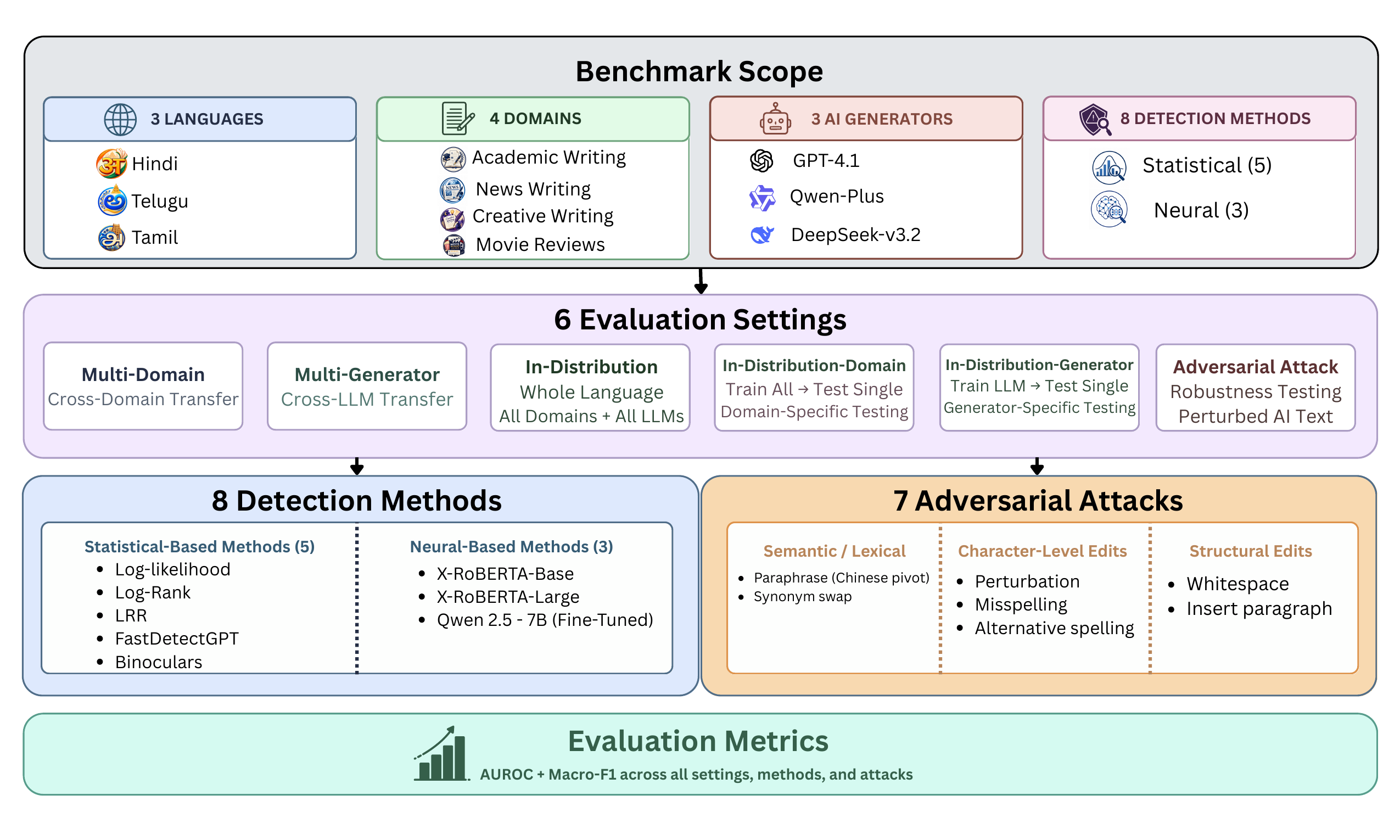}
    \vspace{-6pt}
    \caption{Overview of \textbf{\textit{IndicDetect}}. We curate human-written
texts in three Indic languages (Hindi, Telugu, and Tamil) across four
domains and pair them with LLM-generated counterparts from GPT-4.1,
Qwen-Plus, and DeepSeek-v3.2. Detectors are evaluated under six settings:
In-Distribution, In-Distribution-Domain, In-Distribution-Generator,
Multi-Domain, Multi-Generator, and Multi-Attack. To probe robustness, we
apply realistic adversarial attacks organized into three families: semantic
and lexical rewrites (paraphrasing and synonym substitution), character-level
edits (perturbation, misspelling, and alternative spelling), and structural
edits (whitespace and paragraph insertion). Performance is reported using
AUROC and Macro-F1 across eight detectors spanning statistical methods
(Log-Likelihood, Log-Rank, LRR, Fast-DetectGPT, and Binoculars) and neural
methods (XLM-RoBERTa-Base, XLM-RoBERTa-Large, and a fine-tuned
Qwen~2.5-7B).}
    \label{fig:IndicDetect}
\end{figure*}
\section{IndicDetect}

\textbf{\textit{IndicDetect}} is a comprehensive benchmark designed to evaluate
LLM-generated text detection in Hindi, Telugu, and Tamil
\citep{goyal-etal-2022-flores} under realistic and challenging conditions.
It is built on a multi-domain, multi-generator, multi-task, multi-detector
evaluation pipeline (Figure~\ref{fig:IndicDetect}), comprising carefully
screened human-written texts across four domains (academic, news, creative,
and movie reviews) and texts generated by multiple LLMs using controlled inputs.
All samples undergo identical preprocessing, tokenization, fixed splits, and
uniform scoring to ensure reproducibility, with AUROC and Macro-F1 as the primary
evaluation metrics. The benchmark evaluates detectors under six task settings:
\textit{In-Distribution} (In-Dist.), \textit{In-Distribution Domain}
(In-Dist.(Dom.)), \textit{In-Distribution Generator} (In-Dist.(Gen.)),
\textit{Multi-Domain} (M-Domain), \textit{Multi-Generator} (M-Generator),
and \textit{Multi-Attack} (M-Attack), enabling systematic testing of
generalization across domains, LLMs, and adversarial scenarios. Robustness
is further challenged through meaning-preserving adversarial attacks such as
paraphrasing, perturbations, and character-level noise. \textbf{\textit{IndicDetect}}
evaluates both zero-shot statistical detectors (e.g., Log-Likelihood,
Log-Rank, LRR, Fast-DetectGPT \citep{bao2024fastdetectgpt}, Binoculars
\citep{DBLP:conf/icml/HansSCKSGGG24}) and supervised neural detectors
(e.g., XLM-RoBERTa Base/Large \citep{radford2019gpt2detector}), reporting
results per language and aggregated across all evaluation settings. By
leveraging monolingual, script-constant data and realistic stress tests,
\textbf{\textit{IndicDetect}} addresses critical gaps in Indic AI-generated text
detection and provides a strong foundation for language-aware and robust
detector evaluation and benchmarking. 

\subsection{Dataset Construction}

The \textbf{\textit{IndicDetect}} benchmark is constructed from a carefully curated
collection of human-written and LLM-generated texts in Hindi, Telugu, and
Tamil \citep{kunchukuttan2020ai4bharatindicnlpcorpusmonolingualcorpora}. It is designed to represent real-world language use across multiple
domains while providing strict licensing and reproducibility guarantees \citep{dodge-etal-2021-documenting}. All
human-written data are drawn exclusively from freely accessible, clearly
licensed sources \citep{10.1162/tacl_a_00447} and are divided into training and test splits for reproducibility.

\paragraph{Human-Written Data Collection}

For Telugu, human-written documents are collected from contemporary online
sources. These include Google
News\footnote{\scriptsize \url{https://news.google.com/home?hl=te&gl=IN&ceid=IN:te}}
for news articles, 123Telugu\footnote{\scriptsize \url{https://www.123telugu.com/}}
for movie reviews, the Auchithyam Telugu
Journal\footnote{\scriptsize \url{https://auchithyam.com/advanced/latest/index.php}}
for peer-reviewed academic writing, and Telugu Patala
Lyrics\footnote{\scriptsize \url{https://telugupatalalyrics.blogspot.com/}}
for public-domain creative texts.

For Hindi, Google
News\footnote{\scriptsize \url{https://news.google.com/home?hl=hi&gl=IN&ceid=IN:hi}}
serves as the source for news articles, the IJHR
Journal\footnote{\scriptsize \url{https://www.hindijournal.com/}} provides academic
texts, a curated corpus of public-domain poetry
PDFs\footnote{\scriptsize \url{https://www.selfstudys.com/books/ncert-books-pdf/hindi/class-7}}
(comprising \textit{Sharafath Nahi Mili}, \textit{Sahodara Kavya Mala},
\textit{Jeevan Mag}, \textit{Vedanthasaraha}, \textit{Kumar Mukul Poems},
\textit{Navarang Bhaan} by Mahitha Mishra, and \textit{Pandemic Poems})
represents popular creative writing, and Bollywood
Hungama\footnote{\scriptsize \url{https://www.bollywoodhungama.com/}} provides
film-related opinion texts.

For Tamil, human-written texts are drawn from four domain-representative
sources. News articles are collected from major Tamil news
portals,\footnote{\scriptsize Including Dinamalar
(\url{https://www.dinamalar.com/}), Maalaimalar
(\url{https://www.maalaimalar.com/}), The Hindu Tamil
(\url{https://tamil.thehindu.com/}), and BBC Tamil
(\url{https://www.bbc.com/tamil}).}
covering Tamil Nadu, national, and international reporting. Movie reviews are
sourced from dedicated Tamil cinema review
platforms,\footnote{\scriptsize Including Vikatan (\url{https://www.vikatan.com/}),
Galatta (\url{https://www.galatta.com/}), Indiaglitz
(\url{https://www.indiaglitz.com/}), and Behindwoods
(\url{https://behindwoods.com/}).}
yielding opinionated, domain-specific prose. Academic texts are drawn from
the Tamil Virtual
Academy\footnote{\scriptsize \url{https://www.tamilvu.org/}} and Project
Madurai,\footnote{\scriptsize \url{https://www.projectmadurai.org/}} spanning
disciplines such as linguistics, history, science, and philosophy. Finally,
creative writing samples are collected from established Tamil literary
portals,\footnote{\scriptsize Including Kavithaikal
(\url{https://www.kavithaikal.com/}) and Tamilkavithai
(\url{https://www.tamilkavithai.com/}).} representing both contemporary and
classical verse. All Tamil samples are restricted to texts published before
2022 to prevent temporal contamination with the LLM-generated counterparts and ensure validity.

\paragraph {LLM-Generated Data}

LLM-generated text is produced using large language models \citep{brown2020languagemodelsfewshotlearners} under conditions that reflect real-world usage. All models are accessed via their official APIs
or verified local endpoints, and model identifiers, versions, and access paths
are recorded. Across all generators, we fix the decoding temperature to 0.9
and constrain the generation length to 400--450 tokens, while retaining
provider-default values for the remaining decoding parameters. All prompts,
model versions, and generation metadata \citep{dugan-etal-2024-raid}
are logged to support reproducibility. Text generation employs domain-specific prompts designed to produce realistic
stylistic variation \citep{guo-etal-2023-hc3}. Formal, research-style prompts are used for academic
texts; event descriptions or concise summaries are used for news; open-ended
narrative prompts are used for creative texts; and opinion-oriented prompts \citep{wang2024m4multigeneratormultidomainmultilingual}
are used for movie reviews to encourage analytical expression. Variation
across descriptive, analytical, and narrative prompt structures ensures that
the generated corpus reflects the functional and stylistic diversity \citep{ahuja2023megamultilingualevaluationgenerative} of Hindi,
Telugu, and Tamil in practice.

To ensure LLM-generated samples are not trivially distinguishable from human-written text, 
we adopt a dynamic prompt generation strategy. Static, fixed-template prompts introduce 
artificial regularity into generated outputs. For instance, nearly all LLM-generated 
academic samples in Telugu begin with near-identical introductory sentences, allowing 
detectors to exploit surface-level repetition rather than learning genuine human--AI 
distributional differences. Instead, we extract four domain-representative keywords from 
each human-written sample and prompt the LLM to generate a corresponding sample grounded 
in those keywords, semantically anchoring each generated sample to its human-written 
counterpart. The complete prompt pipelines for Hindi, Telugu, and Tamil are provided in 
Tables~\ref{tab:hindi_prompts}, \ref{tab:telugu_prompts}, and \ref{tab:tamil_prompts}. 
This ensures that observed benchmark scores reflect real-world detection difficulty rather 
than an artifact of generation regularity. \looseness = -1

\paragraph{Preprocessing and Quality Control}
All collected documents pass through a standardized preprocessing pipeline
consisting of Unicode normalization, removal of boilerplate and navigation
artifacts, and strict monolinguality filtering \citep{caswell-etal-2020-language} to exclude code-mixed text.
High-quality textual fragments are then extracted and deduplicated across
splits using a combination of $n$-gram overlap \citep{lee-etal-2022-deduplicating} and embedding-based similarity
filtering. Each text sample is annotated with source metadata,
including its URL or document identifier and the date of access, ensuring
auditability \citep{dodge-etal-2021-documenting} and long-term reproducibility.
Specifically, all texts undergo NFC-level Unicode normalization across Hindi
(Devanagari), Telugu, and Tamil scripts before further processing.
Boilerplate and navigation artifacts are removed using regex-based heuristics
targeting menu strings, pagination tokens, and URL fragments in scraped HTML
content. Samples with more than 5\% of tokens outside the target script Unicode
block are discarded to enforce monolinguality. Deduplication is performed using
5-gram Jaccard overlap (threshold 0.7), followed by multilingual embedding
cosine similarity (threshold 0.85), applied separately across splits to prevent
data leakage. \looseness=-1

\paragraph {Adversarial Attack Generation}

To evaluate detector robustness under realistic conditions, all LLM-generated 
samples are subjected to adversarial attacks designed to mimic the kinds of 
surface-level modifications a user might apply to evade detection. Unless 
otherwise specified, all attacks are carefully designed to preserve the 
original semantics and gold labels, ensuring that the detection task itself 
remains valid. These attacks specifically target the orthographic and 
morphological properties of Brahmic scripts, making them particularly 
relevant for Hindi, Telugu, and Tamil. A detailed analysis of each attack 
type is presented in Section~\ref{sssec:Attacks}.

\paragraph{Paraphrase Attacks} are carried out via back-translation
(Original$\rightarrow$Chinese$\rightarrow$Original) to produce
semantically equivalent rewrites \cite{sennrich-etal-2016-improving}. Natural lexical and syntactic variation is
introduced while preserving meaning, thereby challenging detectors that rely
on local likelihood or textual-stability indicators in robust ways.

\paragraph{Perturbation Attacks} apply random character-level deletions (at a rate of 50\% for all languages). These perturbations respect script-specific grapheme-cluster structure, introducing a small amount of noise that disrupts \citep{dugan-etal-2024-raid}
tokenization while having minimal impact on readability or human comprehension across diverse linguistic contexts in most cases.

\paragraph{Whitespace Addition Attacks} introduce random whitespace insertions \citep{dugan-etal-2024-raid}
and deletions (approximately 20\%) to break subword and token boundaries.
These target tokenizer-sensitive likelihood cues while maintaining human readability and text coherence across languages.

\paragraph{Insert-Paragraph Attacks} distort local coherence by injecting \citep{dugan-etal-2024-raid}
neutral paragraphs drawn from the same source domain, while leaving the
original label unchanged. This probes detectors that rely on surface
continuity or short-range discourse cues without altering meaning. 

\paragraph{Alternative Spelling Attacks} substitute tokens with
orthographically equivalent forms drawn from curated, language-specific
dictionaries \citep{dugan-etal-2024-raid} (e.g., optional diacritics and script-internal alternations).
These replacements preserve meaning while introducing realistic orthographic
variation and typographic diversity.  

\paragraph{Misspelling Attacks} introduce low-rate, realistic typographical
errors using manually compiled, language-specific dictionaries (e.g.,
keyboard-proximity errors and common character omissions). These replacements \citep{dugan-etal-2024-raid}
do not compromise readability but disrupt likelihood- and rank-based detection
signals quite effectively.  

\paragraph{Synonym Swap Attacks} substitute content words with
POS- and lemma-balanced near-synonyms, leaving sentence meaning intact while \citep{dugan-etal-2024-raid}
varying lexical choice. Substitution dictionaries are built from Hindi,
Telugu, and Tamil lexical resources containing human-annotated synonym pairs,
probing detectors that exploit surface-level word preferences.  

\paragraph{Attack Dictionary Construction.}
To build the dictionaries required for the Alternative Spelling, Misspelling,
and Synonym Swap attacks, we leveraged a large language model to automate and
scale the construction process across all three languages. Rather than
manually crafting each entry, we first extracted the most frequent and
linguistically representative keywords from the dataset samples, ensuring
that the dictionaries remained grounded in the actual language patterns
present in the data. These keywords were then used to prompt the LLM to
generate comprehensive substitution lists \citep{dugan-etal-2024-raid} for each attack type. For
Alternative Spelling, the model produced culturally and regionally varied
spellings of the same terms. For Misspelling, it generated plausible
typographical and phonetic errors that a human writer might realistically
make. For Synonym Swap, the model provided semantically equivalent
substitutions that preserved the original meaning while altering the surface
form of the text. This approach enabled us to construct rich, diverse, and
linguistically coherent dictionaries at a scale that would have been
impractical to achieve through manual annotation alone. With these attacks defined, we describe the six evaluation settings in
Section~\ref{sssec:multiattack}.

\paragraph {Corpus Statistics and Evaluation Splits}

\begin{table*}[t]
\centering
\small
\renewcommand{\arraystretch}{0.95}
\setlength{\tabcolsep}{8pt}
\begin{tabular}{llcccccc}
\toprule
\textbf{Lang.} & \textbf{Domain} & \textbf{Human} & \textbf{GPT-4.1}
& \textbf{Qwen-Plus} & \textbf{DS-v3.2} & \textbf{Adv.} & \textbf{Total} \\
\midrule
\multirow{5}{*}{Telugu}
  & Academic   & 1,000 & 1,000 & 1,000 & 1,000 & 3,000 & 7,000 \\
  & News       & 1,000 & 1,000 & 1,000 & 1,000 & 3,000 & 7,000 \\
  & Creative   & 1,000 & 1,000 & 1,000 & 1,000 & 3,000 & 7,000 \\
  & Movie Rev. & 1,000 & 1,000 & 1,000 & 1,000 & 3,000 & 7,000 \\
\cmidrule(lr){2-8}
  & \textit{Subtotal} & \textit{4,000} & \textit{4,000} & \textit{4,000}
  & \textit{4,000} & \textit{12,000} & \textit{28,000} \\
\midrule
\multirow{5}{*}{Hindi}
  & Academic   & 1,000 & 1,000 & 1,000 & 1,000 & 3,000 & 7,000 \\
  & News       & 1,000 & 1,000 & 1,000 & 1,000 & 3,000 & 7,000 \\
  & Creative   & 1,000 & 1,000 & 1,000 & 1,000 & 3,000 & 7,000 \\
  & Movie Rev. & 1,000 & 1,000 & 1,000 & 1,000 & 3,000 & 7,000 \\
\cmidrule(lr){2-8}
  & \textit{Subtotal} & \textit{4,000} & \textit{4,000} & \textit{4,000}
  & \textit{4,000} & \textit{12,000} & \textit{28,000} \\
\midrule
\multirow{5}{*}{Tamil}
  & Academic   & 1,000 & 1,000 & 1,000 & 1,000 & 3,000 & 7,000 \\
  & News       & 1,000 & 1,000 & 1,000 & 1,000 & 3,000 & 7,000 \\
  & Creative   & 1,000 & 1,000 & 1,000 & 1,000 & 3,000 & 7,000 \\
  & Movie Rev. & 1,000 & 1,000 & 1,000 & 1,000 & 3,000 & 7,000 \\
\cmidrule(lr){2-8}
  & \textit{Subtotal} & \textit{4,000} & \textit{4,000} & \textit{4,000}
  & \textit{4,000} & \textit{12,000} & \textit{28,000} \\
\midrule
\multicolumn{2}{l}{\textbf{Grand Total}}
  & \textbf{12,000} & \textbf{12,000} & \textbf{12,000} & \textbf{12,000}
  & \textbf{36,000} & \textbf{84,000} \\
\bottomrule
\end{tabular}
\caption{\textbf{\textit{IndicDetect}} dataset statistics. Adv.\ = adversarial samples
generated by attacking LLM outputs (1,000 per LLM $\times$ 3 LLMs = 3,000 per
domain). DS-v3.2 = DeepSeek-v3.2.}
\label{tab:indicdetect_stats}
\end{table*}

The resulting corpus contains approximately 23.50 million tokens across the
three languages, as summarised in Table~\ref{tab:indicdetect_stats}. For each
evaluation setting, standardised splits of 500 training and 500 test samples
ensure balanced and comparable evaluation \citep{gorman-bedrick-2019-need}
across detectors, with all samples containing at least 400 tokens to minimise
variance from short or fragmentary inputs. Results are reported per language
and macro-averaged over domains, generators, and attack types
\citep{10.5555/3524938.3525348}. To further validate corpus quality, we
conduct a linguistic feature analysis using IndicBERT
\citep{doddapaneni-etal-2023-towards}; full results are presented in
Table~\ref{tab:full_indic_baseline}.\looseness=-1

\subsection{Experiment Settings}

We evaluate detectors under six complementary settings, each targeting a
distinct generalization aspect. To stress-test robustness beyond clean
generations, we incorporate seven adversarial attacks
\citep{dugan-etal-2024-raid} applied exclusively at test time, each adapted
for Hindi, Telugu, and Tamil scripts. 

\paragraph{In-Distribution}

This setting measures detector performance under perfectly matched train-test 
conditions, providing an upper-bound estimate prior to any distribution shift \citep{dugan-etal-2024-raid}. 
Supervised detectors are trained on the ID-train split, while zero-shot 
detectors calibrate only the decision threshold on the development split. 
Stratified splits balance topics, prompts, models, and lengths across all experimental conditions. 

 \begin{figure}[htbp]
\centering
\includegraphics[width=1.00\columnwidth]{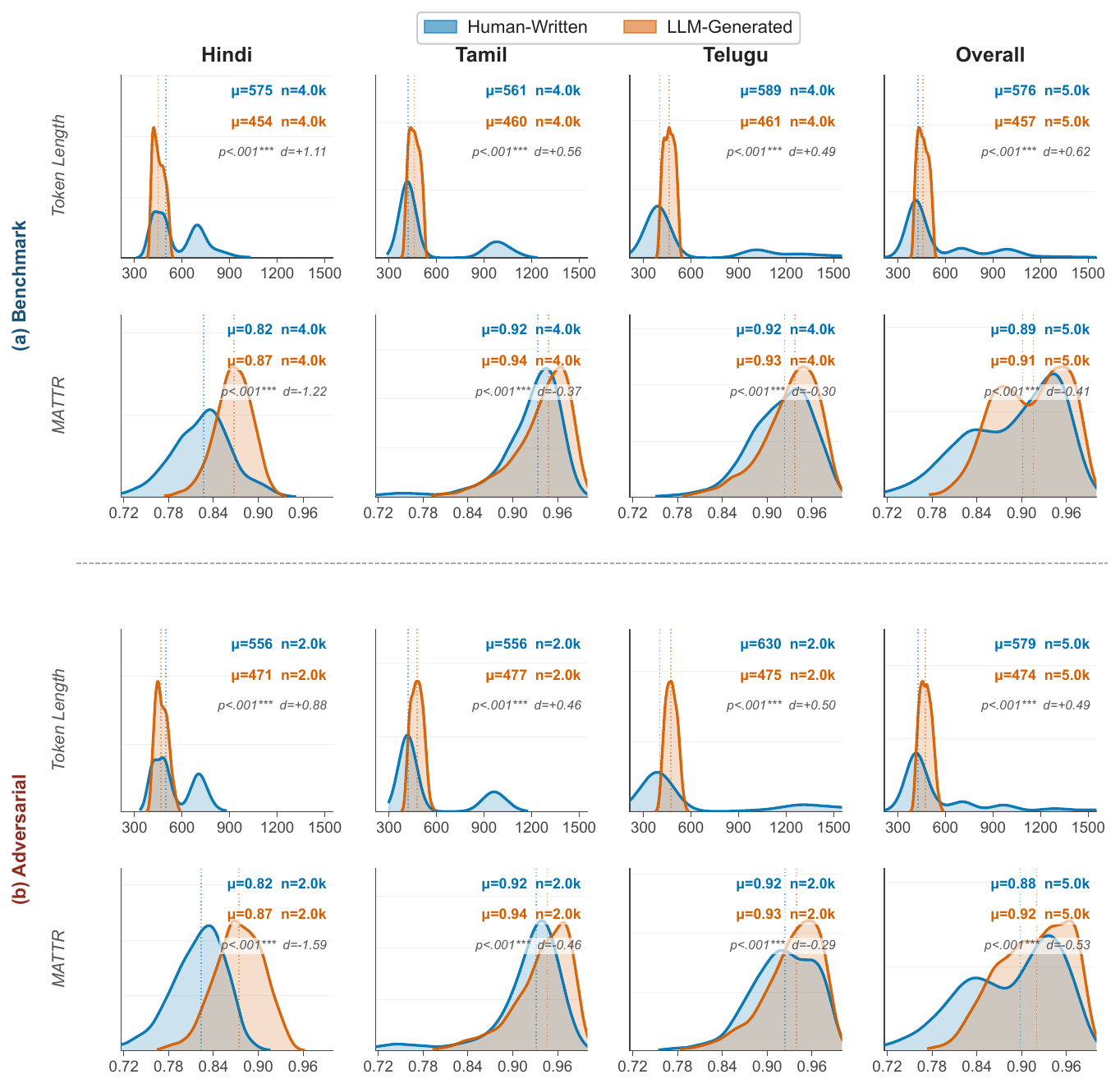}
    \caption{\small XLM-RoBERTa token length and MATTR ($w{=}50$) 
    distributions for human vs.\ LLM texts in Hindi, Tamil, and 
    Telugu. \textbf{(A)}~Benchmark data. \textbf{(B)}~Adversarial 
    attack data.}
    \label{fig:benchmark_stats}
    \vspace{-8pt}
\end{figure}

\paragraph{In-Distribution Domain}
This setting isolates domain sensitivity by evaluating each domain separately \citep{guo-etal-2023-hc3}
(news, academic, review, and creative), revealing detector generalization 
across domains. Supervised detectors are trained on all domains for a given 
language, while zero-shot detectors calibrate only the decision threshold on 
the development split. Performance differences quantify within-language domain 
sensitivity for Hindi, Telugu, and Tamil quite distinctly. 

\paragraph{In-Distribution Generator}

This setting isolates generator sensitivity by evaluating each LLM separately \citep{uchendu2021turingbenchbenchmarkenvironmentturing}
using identical prompts and domains, revealing how well detectors generalize 
across generators without confounding fitting or calibration effects. 
Supervised detectors are trained on all samples for a given language, while 
zero-shot detectors calibrate only the decision threshold on the development 
split. Performance differences across generators quantify generator sensitivity and transferability. 

\paragraph{Multi-Domain}

This setting measures detector generalization beyond domain-specific cues, 
which is critical for real-world deployment \citep{li2023magemaskedgenerativeencoder} where training and test domains 
rarely match. Supervised detectors are trained on one source domain and 
evaluated on the remaining domains, while zero-shot detectors apply the 
source-domain threshold unchanged to target domains. Each domain serves as 
the source in turn and results are macro-averaged subsequently. 

\paragraph{Multi-Generator}

This setting evaluates detector generalization across unseen generators, 
essential since new LLMs emerge after detector training. Supervised detectors \citep{dugan-etal-2024-raid}
are trained on one LLM and evaluated on others under identical prompts and 
domains, while zero-shot detectors apply the source threshold unchanged. 
Human data and length controls are held constant.  

 \begin{figure}[htbp]
\centering
\includegraphics[width=1.05\columnwidth]{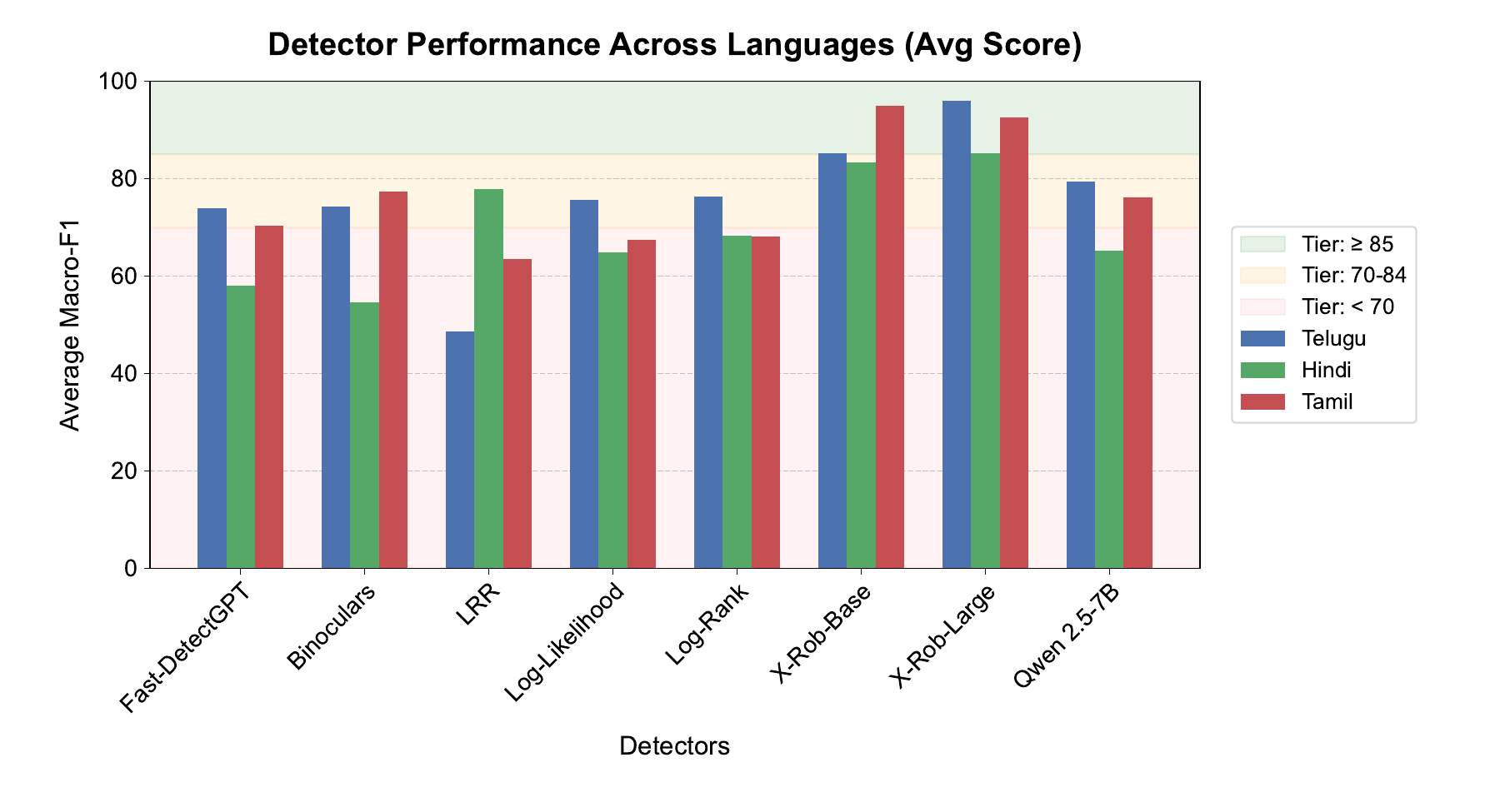}
\caption{Comparison of detector generalization performance (Avg.\ Macro-F1) across Telugu, Hindi, and Tamil. Shaded regions indicate performance tiers: Tier~1 ($\ge 85$), Tier~2 ($70\text{--}84$), and Tier~3 ($<70$). Fine-tuned X-Rob detectors achieve the strongest overall performance, while LRR shows the greatest cross-language inconsistency.}
    \label{fig:detector_performance}
    \vspace{-8pt}
\end{figure}

\paragraph {Multi-Attack} \label{sssec:multiattack}

This setting measures detector robustness to surface-level perturbations, 
critical for real-world deployment \citep{dugan-etal-2024-raid} where text is often edited to evade 
detection. Detectors are trained on clean human-LLM pairs and evaluated 
on attacked test sets across Hindi, Telugu, and Tamil under diverse conditions. Detailed experimental setup is provided in Appendix~\ref{sssec:Attacks}.

\paragraph{Generalization Score}

To summarize out-of-distribution performance in a single comparable figure,
we define a generalization score $G_d$ for detector $d$ as the mean Macro-F1
over the $N$ out-of-distribution evaluation conditions:
\begin{equation}
    G_d = \frac{1}{N} \sum_{i=1}^{N} S_{d,i},
    \qquad N = 3,
\end{equation}
where $S_{d,i}$ is the Macro-F1 of detector $d$ under the $i$-th
out-of-distribution condition. The conditions are Multi-Domain,
Multi-Generator, and Multi-Attack, each with uniform weight. Each $S_{d,i}$
is itself macro-averaged over its sub-conditions: four source domains, three
generators, and seven attack types, respectively. In-distribution settings
are excluded, as they measure matched-condition performance, and AUROC is
reported for reference only and does not enter $G_d$. This scalar metric
enables direct comparison of detectors across Hindi, Telugu, and Tamil under
realistic distribution shifts.\looseness=-1

\begin{table*}[t]
\centering
\small
\renewcommand{\arraystretch}{0.82}
\setlength{\tabcolsep}{3pt}
\resizebox{\textwidth}{!}{%
\begin{tabular}{lcccccccccccccccc}
\toprule
& \multicolumn{2}{c}{\textbf{M-Domain}}
& \multicolumn{2}{c}{\textbf{M-Generator}}
& \multicolumn{2}{c}{\textbf{M-Attack}}
& \multicolumn{2}{c}{\textbf{In-Dist.}}
& \multicolumn{2}{c}{\textbf{In-Dist.(Dom.)}}
& \multicolumn{2}{c}{\textbf{In-Dist.(Gen.)}}
& \multicolumn{3}{c}{\textbf{Generalization}}
& \\
\cmidrule(lr){2-3}\cmidrule(lr){4-5}\cmidrule(lr){6-7}
\cmidrule(lr){8-9}\cmidrule(lr){10-11}\cmidrule(lr){12-13}
\cmidrule(lr){14-16}
\textbf{Detector}
& AUC & $F_1$ & AUC & $F_1$ & AUC & $F_1$
& AUC & $F_1$ & AUC & $F_1$ & AUC & $F_1$
& Dom & Gen & Atk
& \textbf{Avg} \\
\midrule
\multicolumn{17}{l}{\textit{Telugu}} \\
\midrule
Fast-DetectGPT & 83.38 & 80.60 & 83.39 & 76.01 & 72.21 & 65.32 & 83.47 & 77.88 & 83.38 & 80.23 & 72.86 & 57.80 & 80.60 & 76.01 & 65.32 & \md{73.98} \\
Binoculars     & 86.75 & 82.60 & 86.16 & 79.67 & 72.39 & 60.65 & 86.22 & 81.82 & 86.76 & 81.15 & 79.65 & 65.14 & 82.60 & 79.67 & 60.65 & \md{74.31} \\
LRR            & 56.51 & 47.26 & 55.25 & 41.67 & 62.36 & 56.87 & 55.13 & 54.46 & 56.51 & 43.45 & 63.30 & 46.35 & 47.26 & 41.67 & 56.87 & \lo{48.60} \\
Log-Likelihood & 94.04 & 88.06 & 84.89 & 78.10 & 74.62 & 60.77 & 84.90 & 82.35 & 94.04 & 78.50 & 80.47 & 52.24 & 88.06 & 78.10 & 60.77 & \md{75.64} \\
Log-Rank       & 93.92 & 88.48 & 83.55 & 76.54 & 74.82 & 63.91 & 83.56 & 80.89 & 93.91 & 74.68 & 80.44 & 53.28 & 88.48 & 76.54 & 63.91 & \md{76.31} \\
X-Rob-Base     & 99.88 & 89.70 & 80.84 & 69.25 & 99.98 & 96.77 & 99.99 & 95.45 & 99.98 & 97.73 & 94.76 & 79.03 & 89.70 & 69.25 & 96.77 & \hi{\secbest{85.24}} \\
X-Rob-Large    & 99.99 & 95.70 & 99.99 & 96.25 & 99.99 & 96.22 & 99.99 & 97.68 & 99.99 & 93.03 & 99.99 & 84.98 & 95.70 & 96.25 & 96.22 & \hi{\best{96.06}} \\
Qwen 2.5-7B    & 63.62 & 55.15 & 93.01 & 86.59 & 99.38 & 96.24 & 99.86 & 98.37 & 99.58 & 99.03 & 91.79 & 87.44 & 55.15 & 86.59 & 96.24 & \md{79.33} \\
\midrule
\multicolumn{17}{l}{\textit{Hindi}} \\
\midrule
Fast-DetectGPT & 63.39 & 76.31 & 61.84 & 59.36 & 41.82 & 38.54 & 61.84 & 64.44 & 61.40 & 64.99 & 61.84 & 57.40 & 76.31 & 59.36 & 38.54 & \lo{58.07} \\
Binoculars     & 75.98 & 68.28 & 76.03 & 68.13 & 35.71 & 27.39 & 76.04 & 73.33 & 75.97 & 72.60 & 76.03 & 64.94 & 68.28 & 68.13 & 27.39 & \lo{54.60} \\
LRR            & 84.00 & 87.17 & 77.65 & 75.46 & 66.04 & 70.88 & 77.68 & 86.04 & 84.03 & 85.89 & 77.65 & 75.07 & 87.17 & 75.46 & 70.88 & \md{77.84} \\
Log-Likelihood & 88.73 & 89.71 & 75.57 & 69.12 & 48.70 & 35.82 & 75.57 & 77.72 & 88.73 & 69.45 & 77.65 & 75.07 & 89.71 & 69.12 & 35.82 & \lo{64.88} \\
Log-Rank       & 91.23 & 92.72 & 76.96 & 70.96 & 53.24 & 41.06 & 76.97 & 78.43 & 91.22 & 69.44 & 76.96 & 70.19 & 92.72 & 70.96 & 41.06 & \lo{68.25} \\
X-Rob-Base     & 99.94 & 95.13 & 79.06 & 88.57 & 96.98 & 66.21 & 99.99 & 92.40 & 99.97 & 96.59 & 99.98 & 98.39 & 95.13 & 88.57 & 66.21 & \md{\secbest{83.30}} \\
X-Rob-Large    & 100.00 & 97.04 & 99.99 & 97.15 & 99.37 & 61.32 & 99.99 & 96.02 & 99.99 & 92.61 & 99.99 & 96.12 & 97.04 & 97.15 & 61.32 & \hi{\best{85.17}} \\
Qwen 2.5-7B    & 66.94 & 57.90 & 85.67 & 77.78 & 94.00 & 60.18 & 99.95 & 99.20 & 99.95 & 98.75 & 99.93 & 98.92 & 57.90 & 77.78 & 60.18 & \lo{65.29} \\
\midrule
\multicolumn{17}{l}{\textit{Tamil}} \\
\midrule
Fast-DetectGPT & 84.31 & 78.32 & 84.08 & 76.59 & 63.83 & 55.96 & 84.11 & 81.38 & 84.31 & 80.40 & 84.08 & 74.80 & 78.32 & 76.59 & 55.96 & \md{70.29} \\
Binoculars     & 90.29 & 86.51 & 89.94 & 83.26 & 69.77 & 62.33 & 89.92 & 86.42 & 90.30 & 86.51 & 89.94 & 81.36 & 86.51 & 83.26 & 62.33 & \md{77.37} \\
LRR            & 45.66 & 61.18 & 47.05 & 43.05 & 56.09 & 86.06 & 47.13 & 86.16 & 45.59 & 86.06 & 47.05 & 67.47 & 61.18 & 43.05 & 86.06 & \lo{63.43} \\
Log-Likelihood & 82.09 & 77.83 & 80.35 & 74.09 & 60.99 & 50.22 & 80.35 & 79.52 & 82.09 & 76.30 & 80.35 & 72.49 & 77.83 & 74.09 & 50.22 & \lo{67.38} \\
Log-Rank       & 80.16 & 76.45 & 77.91 & 71.74 & 62.67 & 56.24 & 77.91 & 79.98 & 80.18 & 77.33 & 77.91 & 72.08 & 76.45 & 71.74 & 56.24 & \lo{68.14} \\
X-Rob-Base     & 99.96 & 93.61 & 99.94 & 98.16 & 99.96 & 93.17 & 99.99 & 95.55 & 99.97 & 89.17 & 99.98 & 96.53 & 93.61 & 98.16 & 93.17 & \hi{\best{94.98}} \\
X-Rob-Large    & 99.99 & 97.59 & 99.90 & 97.60 & 99.99 & 82.46 & 99.99 & 94.15 & 99.99 & 89.60 & 99.99 & 98.49 & 97.59 & 97.60 & 82.46 & \hi{\secbest{92.55}} \\
Qwen 2.5-7B    & 62.90 & 57.81 & 88.42 & 81.22 & 98.73 & 89.31 & 99.41 & 96.33 & 96.85 & 96.10 & 99.58 & 97.36 & 57.81 & 81.22 & 89.31 & \md{76.11} \\
\bottomrule
\end{tabular}%
}
\caption{Detector leaderboard across Telugu, Hindi, and Tamil.
M- represents Multi-, AUC represents AUROC, ${F_1}$ is the Macro-F1, and
Dom/Gen/Atk are the task-setting scores. \textbf{Avg} is colour-coded by tier:
\setlength{\fboxsep}{2pt}\colorbox{avgHigh}{${\geq}85$},
\setlength{\fboxsep}{2pt}\colorbox{avgMid}{$70$--$84$}, and
\setlength{\fboxsep}{2pt}\colorbox{avgLow}{${<}70$}. Within each tier,
\textbf{\underline{bold+underline}} = 1st and
\textbf{\textit{bold+italic}} = 2nd best Avg per language.}
\label{tab:leaderboard_unified}
\end{table*}

\section{Detectors}
We evaluate eight detectors spanning training-free and supervised approaches.
Training-free methods include Log-Likelihood \citep{DBLP:journals/corr/abs-1908-09203},
Log-Rank \citep{su2023detectllm}, LRR \citep{su2023detectllm},
Fast-DetectGPT \citep{bao2024fastdetectgpt}, and Binoculars
\citep{DBLP:conf/icml/HansSCKSGGG24}. Supervised baselines are
XLM-RoBERTa-Base and XLM-RoBERTa-Large \citep{radford2019gpt2detector},
fine-tuned on labeled human-machine pairs, and Qwen~2.5-7B
\citep{qwen2025qwen25technicalreport} after task-specific Indic fine-tuning.
All detectors share a unified preprocessing pipeline with identical
train-validation-test splits where F1 thresholds are selected on held-out
validation data and threshold-free metrics are additionally reported for comprehensive evaluation.

\section{Results and Discussion}

\paragraph{Main Results}
XLM-RoBERTa-Large \citep{radford2019gpt2detector} achieves the highest average score in Telugu (96.06) and Hindi (85.17) and ranks a close second in Tamil (92.55), where XLM-RoBERTa-Base \citep{radford2019gpt2detector} leads (94.98). XLM-RoBERTa-Base \citep{radford2019gpt2detector} is otherwise the consistent runner-up, in Telugu (85.24) and Hindi (83.30). The fine-tuned Qwen~2.5-7B \citep{qwen2025qwen25technicalreport} achieves an average score of 79.33 in Telugu but trails the best zero-shot method in Hindi (65.29 vs.\ LRR 77.84) and Tamil (76.11 vs.\ Binoculars \citep{DBLP:conf/icml/HansSCKSGGG24} 77.37). Among zero-shot methods, Log-Rank \citep{su2023detectllm} leads in Telugu (76.31), LRR \citep{su2023detectllm} in Hindi (77.84), and Binoculars \citep{DBLP:conf/icml/HansSCKSGGG24} in Tamil (77.37). Fast-DetectGPT \citep{bao2024fastdetectgpt} and likelihood-based methods remain middling across settings, while LRR is highly inconsistent, collapsing in Telugu (48.60) and Tamil (63.43). These per-language patterns across all six settings are visualized in Figure~\ref{fig:radar}. \looseness=-2

\paragraph{Domain-Robustness}
XLM-RoBERTa-Large \citep{radford2019gpt2detector} maintains the strongest cross-domain performance across all languages (Multi-Domain $F_1$ of 95.70, 97.04, and 97.59 in Telugu, Hindi, and Tamil), with XLM-RoBERTa-Base \citep{radford2019gpt2detector} close behind. In contrast, Qwen~2.5-7B \citep{qwen2025qwen25technicalreport} is notably weak under domain shift (Multi-Domain $F_1$ of 55.15, 57.90, and 57.81), indicating limited domain adaptation despite its stronger performance under other evaluation settings. Likelihood-based methods \citep{DBLP:journals/corr/abs-1908-09203, su2023detectllm} are surprisingly domain-robust in Telugu (88.06, 88.48) and Hindi (89.71, 92.72) but degrade on Tamil, while LRR \citep{su2023detectllm} collapses on Telugu (47.26) and Tamil (61.18) domain-shifted text. 

\paragraph{Generator-Robustness}
XLM-RoBERTa-Large \citep{radford2019gpt2detector} generalizes best to unseen generators with minimal degradation (Multi-Generator $F_1$ of 96.25, 97.15, and 97.60). Qwen~2.5-7B \citep{qwen2025qwen25technicalreport} also transfers strongly across generators (86.59, 77.78, 81.22), whereas XLM-RoBERTa-Base \citep{radford2019gpt2detector} is highly variable, weak in Telugu (69.25) yet strong in Hindi (88.57) and Tamil (98.16). Fast-DetectGPT \citep{bao2024fastdetectgpt} and Binoculars \citep{DBLP:conf/icml/HansSCKSGGG24} drop noticeably on unseen generators, most severely in Hindi (59.36 and 68.13), and LRR \citep{su2023detectllm} degrades sharply in Telugu (41.67) and Tamil (43.05).

\paragraph{Attack-Robustness}
XLM-RoBERTa-Large \citep{radford2019gpt2detector} shows remarkable resilience, maintaining M-Attack AUROC above 99.37 across all languages. Qwen~2.5-7B \citep{qwen2025qwen25technicalreport} is also highly robust, with Multi-Attack AUROC above 94. Zero-shot methods suffer most: LRR \citep{su2023detectllm} exhibits severe degradation in Tamil (AUROC 56.09), and likelihood- and curvature-based methods collapse under paraphrasing and insertion attacks, particularly in Hindi (Multi-Attack $F_1$ as low as 27.39 for Binoculars and 35.82 for Log-Likelihood \citep{DBLP:journals/corr/abs-1908-09203}). Hindi proves the most vulnerable language overall, where even fine-tuned models see sharp $F_1$ drops (XLM-RoBERTa-Large \citep{radford2019gpt2detector} 61.32, XLM-RoBERTa-Base 66.21 \citep{radford2019gpt2detector}), underscoring the need for adversarially aware training in practice.

\section{Conclusion}
We presented \textbf{\textit{IndicDetect}}, a carefully constructed benchmark for
evaluating LLM-generated text detection in Hindi, Telugu, and Tamil under
realistic conditions. The benchmark standardizes data collection, splitting,
and evaluation, and systematically examines detector generalization under
domain shifts, generator shifts, and adversarial attacks across seven attack
types targeting Brahmic script properties. We find that supervised neural
detectors consistently outperform zero-shot alternatives: fine-tuned
XLM-RoBERTa models \citep{radford2019gpt2detector} achieve the best average
performance across all three languages, with XLM-RoBERTa-Large \citep{radford2019gpt2detector} the strongest
overall and XLM-RoBERTa-Base \citep{radford2019gpt2detector} a close runner-up. Qwen~2.5-7B
\citep{qwen2025qwen25technicalreport} attains strong overall scores and
remains highly robust to unseen generators and adversarial attacks, yet
degrades sharply under domain shift, revealing limited domain adaptation
despite its high ranking. In contrast, zero-shot methods such as
Log-Likelihood \citep{DBLP:journals/corr/abs-1908-09203},
Log-Rank \citep{su2023detectllm}, and
Binoculars \citep{DBLP:conf/icml/HansSCKSGGG24} degrade rapidly under unseen
generators and adversarial perturbations, and LRR \citep{su2023detectllm}
is especially unstable, collapsing on Telugu and Tamil. Across languages,
Hindi proves the most challenging setting, exhibiting the lowest average
scores and the greatest vulnerability to adversarial attacks, where even
fine-tuned detectors suffer sharp performance drops, underscoring the need
for adversarially aware training for Indic languages.

\section*{Acknowledgments}

This work was supported in part by the Science and Technology Development Fund of Macau SAR (Grant Nos.\ FDCT/0007/2024/AKP, EF2024-00185-FST), the UM and UMDF (Grant Nos.\ MYRG-GRG2024-00165-FST-UMDF, MYRG-GRG2025-00236-FST), the Tencent AI Lab Rhino-Bird Research Program (Grant No.\ EF2023-00151-FST), the Stanley Ho Medical Development Foundation (Grant No.\ SHMDF-AI/2026/001), and the National Natural Science Foundation of China (Grant No.\ 62266013).

\section*{Limitations}
\textbf{\textit{IndicDetect}} covers Hindi, Telugu, and Tamil only; broader Indic
language coverage was precluded by the absence of large-scale, domain-balanced
corpora, and computational constraints limited evaluation to smaller models.
Moreover, the rapidly evolving LLM landscape means newer generators may not be
represented, and our seven attack types do not exhaust possible evasion
strategies such as adaptive attacks or human post-editing. Finally, our
pipeline-bound data and binary human/machine labels may not fully capture
in-the-wild text, including code-mixed or transliterated content and
partially edited or co-authored documents.

\bibliography{latex/anthology} 

\newpage
\appendix
\section{Background}
\subsection{Computational Experiments}
The high-performance computing cluster (HPCC) at our university was used to run all experiments on NVIDIA A100 GPUs with 80GB of memory. The large GPU memory enabled efficient large-scale evaluation across many detectors, domains, generators, and adversarial environments, and stabilized training and inference without resource-induced bottlenecks. Such an arrangement promotes the reproducibility and scalability of the reported experimental results. For reproducibility, training hyperparameters are provided in Table~\ref{tab:training_hyperparameters}, and the corresponding reference and scoring models for each detector are detailed in Table~\ref{tab:source_reference_detectors}.

\subsection{Ethics Statement}
In developing IndicDetect, we gather publicly licensed human texts from domains where there is a high risk of abuse and generate corresponding machine texts with commonly used LLMs, then the detectors are evaluated under a carefully planned series of attacks (applied at the testing stage) to test their robustness and simulate real-world conditions. Our aim is to push the detection of LLM-generated text further by offering a transparent, reproducible benchmark which helps to build stronger and more generalizable methods. We acknowledge the dual-use risk: the exposure of a transparent construction pipeline and attack suite could also reveal techniques for evading existing detectors.

 However, we still think that the principle of openness will lead to faster development of detection systems that are less susceptible to being evaded and that responsible use will come along with it. The majority of the data was manually reviewed, but some residual risks remain (e.g., unintentional PII or offensive content); thus, the resource is intended for academic use only, and users should exercise caution.

 \subsection{Data Collection}
For Telugu, human-written documents are collected from contemporary online
sources. These include Google
News\footnote{\scriptsize \url{https://news.google.com/home?hl=te&gl=IN&ceid=IN:te}}
for news articles, 123Telugu\footnote{\scriptsize \url{https://www.123telugu.com/}}
for movie reviews, the Auchithyam Telugu
Journal\footnote{\scriptsize \url{https://auchithyam.com/advanced/latest/index.php}}
for peer-reviewed academic writing, and Telugu Patala
Lyrics\footnote{\scriptsize \url{https://telugupatalalyrics.blogspot.com/}}
for public-domain creative texts.

For Hindi, Google
News\footnote{\scriptsize \url{https://news.google.com/home?hl=hi&gl=IN&ceid=IN:hi}}
serves as the source for news articles, the IJHR
Journal\footnote{\scriptsize \url{https://www.hindijournal.com/}} provides academic
texts, a curated corpus of public-domain poetry
PDFs\footnote{\scriptsize \url{https://www.selfstudys.com/books/ncert-books-pdf/hindi/class-7}}
(comprising \textit{Sharafath Nahi Mili}, \textit{Sahodara Kavya Mala},
\textit{Jeevan Mag}, \textit{Vedanthasaraha}, \textit{Kumar Mukul Poems},
\textit{Navarang Bhaan} by Mahitha Mishra, and \textit{Pandemic Poems})
represents popular creative writing, and Bollywood
Hungama\footnote{\scriptsize \url{https://www.bollywoodhungama.com/}} provides
film-related opinion texts.

For Tamil, human-written texts are drawn from four domain-representative
sources. News articles are collected from major Tamil news
portals,\footnote{\scriptsize Including Dinamalar
(\url{https://www.dinamalar.com/}), Maalaimalar
(\url{https://www.maalaimalar.com/}), The Hindu Tamil
(\url{https://tamil.thehindu.com/}), and BBC Tamil
(\url{https://www.bbc.com/tamil}).}
covering Tamil Nadu, national, and international reporting. Movie reviews are
sourced from dedicated Tamil cinema review
platforms,\footnote{\scriptsize Including Vikatan (\url{https://www.vikatan.com/}),
Galatta (\url{https://www.galatta.com/}), Indiaglitz
(\url{https://www.indiaglitz.com/}), and Behindwoods
(\url{https://behindwoods.com/}).}
yielding opinionated, domain-specific prose. Academic texts are drawn from
the Tamil Virtual
Academy\footnote{\scriptsize \url{https://www.tamilvu.org/}} and Project
Madurai,\footnote{\scriptsize \url{https://www.projectmadurai.org/}} spanning
disciplines such as linguistics, history, science, and philosophy. Finally,
creative writing samples are collected from established Tamil literary
portals,\footnote{\scriptsize Including Kavithaikal
(\url{https://www.kavithaikal.com/}) and Tamilkavithai
(\url{https://www.tamilkavithai.com/}).} representing both contemporary and
classical verse. All Tamil samples are restricted to texts published before
2022 to prevent temporal contamination with the LLM-generated counterparts.

\subsection{Generative models}

We evaluate widely used LLMs that mirror real-world usage. Each model is
accessed via its official API or a local checkpoint and we record the exact model
identifiers and access paths for reproducibility. Across all generators, we fix
the decoding temperature to 0.9 and constrain the generation length to
400--450 tokens, while retaining provider-default values for all remaining
decoding parameters. Prompts and model/version metadata are logged to enable
faithful replication of all generations.

\begin{table}[H]
  \centering
  \small
  \setlength{\tabcolsep}{8pt}
  \begin{tabular}{l l l}
    \toprule
    \textbf{Model} & \textbf{Provider} & \textbf{Access} \\
    \midrule
    GPT\textendash 4.1         & OpenAI        & API \\
    Qwen\textendash Plus & Alibaba Qwen  & API \\
    DeepSeek\textendash V3.2    & DeepSeek      & API \\
    \bottomrule
  \end{tabular}
  \caption{Generative models used to synthesize machine text in our experiments.}
  \label{tab:gen_models}
\end{table}

{GPT-4.1} \citep{blog2022chatgpt} is an API-based model from OpenAI
with strong instruction-following, long-context, and text-generation
capabilities. Its ability to generate fluent and diverse text across
topics and styles makes it a realistic generator for evaluating
cross-model robustness in our benchmark.

{Qwen-Plus} \citep{yang2025qwen3technicalreport} is an instruction-tuned model in Alibaba’s Qwen3 family. It is optimized for fast, low-latency generation while maintaining strong reasoning and multilingual performance (including Indic scripts), making it a practical generator for large-scale benchmarking.

{DeepSeek-V3.2} \citep{deepseekai2025deepseekv32pushingfrontieropen} is a large decoder-only model trained with long-context optimization and efficient sampling strategies. It is known for balanced performance on reasoning, coding, and general dialogue, providing stylistically distinct outputs that are valuable for cross-generator robustness tests.

\subsection{Data Generation settings}
\label{data_generation_settings}

All text generation tasks mentioned in this research were carried out through interactive LLM prompting, employing different prompt structures based on each domain's needs, which led to variety and realism. Across all LLMs, we fixed the decoding temperature to 0.9 and constrained the generation length to 400–450 tokens (min=400, max=450). For academic writing, formal and logical flow was achieved through structured prompts that called forth research-like abstracts. In the case of news writing, the model was given short event cues or concise summaries and was thus expected to turn them into full, coherent, and good quality articles. For creative writing, the use of open-ended and imaginative prompts resulted in story-like passages, rich in narrative. Opinion-oriented prompts were applied in the area of movie reviewing to generate subjective but at the same time analytical thoughts. Thus, varying the formats of the prompts from descriptive to analytical and narrative forms, we came to the conclusion that the output text reflected the full extent of the stylistic and functional variations inherent in the respective real-world language use across domains. For reproducibility, the prompting templates employed for Telugu, Hindi, and Tamil across all domains are provided in Tables~\ref{tab:telugu_prompts}, \ref{tab:hindi_prompts}, and \ref{tab:tamil_prompts}.

\subsection {Attacks} \label{sssec:Attacks}
This section provides an elaborate description of the adversarial attack methods used to assess the detector's robustness to realistic perturbations. Every attack is made to maintain the original semantic content while injecting surface-level or structural perturbations that are generally found in the editing of natural text. Each of the attacks presented here is described according to its motivation, construction procedure, and language-specific considerations, such as paraphrasing, substituting synonyms, misspellings, whitespace and formatting perturbations, and other character-level alterations. These descriptions explain how each attack stresses detector signals while preserving experimental consistency across Hindi, Telugu, and Tamil.

\subsubsection{Paraphrase Attacks}\label{sssec:paraphrase}
Paraphrase attacks are produced using the back-translation method with a set pivot language. For Hindi, Telugu, and Tamil, the samples are translated using the pipeline $\text{Original} \rightarrow \text{Chinese} \rightarrow \text{Original}$  via the official Google Translate API. The process maintains semantic content and introduces lexical and syntactic variation, which is a reproducible process of testing the robustness of detectors to distribution shifts due to paraphrasing. Representative paraphrase samples for Telugu, Hindi, and Tamil, generated using this back-translation pipeline, are illustrated in Tables~\ref{tab:paraphrase_telugu}, \ref{tab:paraphrase_hindi}, and \ref{tab:paraphrase_tamil}.

\subsubsection{Perturbation Attacks} \label{sssec:petrubation}

Perturbation attacks introduce surface-level noise through the random removal of characters in the text. Specifically, in each sample, we independently remove each character with probability $p = 0.5$, while approximately preserving the underlying semantic content. Formally, let a text sample be represented as a sequence of characters
\[
\mathbf{x} = (c_1, c_2, \dots, c_n),
\]
where $c_i$ denotes the $i$-th character. For each character $c_i$, we independently sample a Bernoulli random variable
\[
z_i \sim \text{Bernoulli}(1 - p),
\]
and construct the perturbed text as
\[
\tilde{\mathbf{x}} = \{\, c_i \mid z_i = 1 \,\}.
\]
This corruption of characters' levels emulates natural typing errors and small text corruption, which are typically found in real-world systems, and purposefully corrupts tokenization and local distributional statistics. These perturbations are mostly aimed at detectors sensitive to delicate surface cues, and they are an effective stress test of detectors against the least but still realistic noise. Sample instances of this perturbation attack are shown in Tables~\ref{tab:perturbation_telugu}, \ref{tab:perturbation_hindi}, and \ref{tab:perturbation_tamil}.

\subsubsection{White Space Addition Attacks} \label{sssec:whitespace}
Whitespace attacks aim to add surface-level formatting noise to a text by changing the spacing patterns between textual units, without altering characters or lexical meaning. In contrast with character-level perturbation attacks, no characters or words are changed in the given text, but the length of the existing whitespace segments is chosen selectively in this attack. This leaves the semantic content and readability intact, while deliberately violating the token boundaries and tokenizer behavior.

Formally, let a text sample be represented as an ordered sequence of non-whitespace spans and whitespace segments
\[
\mathbf{x} = (s_1, w_1, s_2, w_2, \dots, s_m),
\]
where each $s_i$ denotes a contiguous non-whitespace span and each $w_i$ denotes a contiguous whitespace segment.~Let $\mathcal{I}=\{i|w_i\text{ is a whitespace segment}\}$ be the index set of whitespace segments, with $|\mathcal{I}|=N_w$. Given a whitespace modification rate $\theta$ (expressed as a percentage), the number of whitespace segments to be modified is defined as
\[
k = \max\left(1, \left\lfloor \frac{\theta}{100} \cdot N_w \right\rfloor \right).
\]

We then sample $k$ indices independently and uniformly from $\mathcal{I}$ with replacement, and for each selected index $i$, the corresponding whitespace segment is modified by appending an additional space character:
\[
w_i \leftarrow w_i \,\Vert\, \text{\texttt{space}} \,.
\]
The resulting whitespace-attacked text is denoted as $\mathbf{x}^{\text{ws}}$ and is obtained by concatenating the modified sequence. This attack primarily affects whitespace-sensitive tokenization and local surface statistics, serving as a targeted stress test for detectors that rely on precise segmentation and formatting cues. Sample instances of the whitespace attack for Telugu, Hindi, and Tamil are presented in Tables~\ref{tab:whitespace_telugu}, \ref{tab:whitespace_hindi}, and \ref{tab:whitespace_tamil}.

\subsubsection{Insert-Paragraph Attacks} \label{sssec:InsertPara}
The Insert-Paragraph attack introduces discourse-level changes to the text's formatting by adding paragraph breaks, without altering its lexical content. Specifically, we first split each sample into sentences (using a language-aware tokenizer when available, with a regular-expression fallback), then select a subset of sentence boundaries and replace the standard inter-sentence space with a paragraph marker (two newline characters). The transformation leaves the words, characters, and sentence order unchanged, but alters the document's layout and local context windows, which can in turn affect detectors sensitive to segmentation, tokenization, and formatting cues. Sample instances of the insert-paragraph attack for Telugu, Hindi, and Tamil are presented in Tables~\ref{tab:insert_paragraph_telugu}, \ref{tab:insert_paragraph_hindi}, and \ref{tab:insert_paragraph_tamil}.

Formally, let a text sample be split into a sequence of sentences
\[
\mathbf{x} = (s_1, s_2, \dots, s_n),
\]
and let the set of candidate insertion positions be the sentence-boundary indices
\[
\mathcal{B} = \{1,2,\dots,n-1\}.
\]
Given an insertion rate $\theta$ (in percent), we choose the number of paragraph breaks as
\[
k = \max\left(1,\ \mathrm{round}\!\left(\frac{\theta}{100}\cdot |\mathcal{B}|\right)\right),
\]
and sample a subset $\mathcal{C} \subseteq \mathcal{B}$ with $|\mathcal{C}| = k$ uniformly without replacement. The insert-paragraph attacked text, denoted as $\mathbf{x}^{\text{ins}}$, is constructed by concatenating sentences with boundary-dependent delimiters:
\[
\mathbf{x}^{\text{ins}} \;=\; s_1 \,\Vert\, d_1 \,\Vert\, s_2 \,\Vert\, d_2 \,\Vert\, \cdots \,\Vert\, d_{n-1} \,\Vert\, s_n,
\]
where
\[
d_i =
\begin{cases}
\texttt{``\textbackslash n\textbackslash n''} & \text{if } i \in \mathcal{C},\\
\texttt{`` ''} & \text{otherwise}.
\end{cases}
\]

\subsubsection{Alternative Spelling Attacks}\label{sssec:Alternativespel}
Alternative Spelling Attacks are aimed at providing realistic orthographic variation to some of the words by substituting them with correct alternative spellings, keeping the original semantic meaning. This attack exploits a manually edited alternative-spelling dictionary for Hindi, Telugu, and Tamil, where every canonical word form is indexed by a set of more frequently occurring spelling variations in real text. These variations are often brought about by dialectal preference, informal writing style, or non-standard transliteration and, as such, form a natural and linguistically reasonable source of distributional shift.
\begin{table}[t]
\centering
\small
\setlength{\tabcolsep}{6pt}
\renewcommand{\arraystretch}{1.2}

\begin{tabular}{l|l}
\toprule
\textbf{Canonical Form} & \textbf{Alternative Spellings} \\
\midrule
\multicolumn{2}{c}{\textbf{Telugu}} \\
\midrule
\foreignlanguage{telugu}{సాహిత్యం} 
& \foreignlanguage{telugu}{సాహిత్యము, సాహిత్యాం} \\

\foreignlanguage{telugu}{పరిణామం} 
& \foreignlanguage{telugu}{పరిణామము, పరిణామం} \\

\foreignlanguage{telugu}{సాంకేతికత} 
& \foreignlanguage{telugu}{సాంకేతికతా, సాంకేతికము} \\

\foreignlanguage{telugu}{అధ్యయనం} 
& \foreignlanguage{telugu}{అధ్యయనము, అధ్యయనం} \\
\midrule
\multicolumn{2}{c}{\textbf{Hindi}} \\
\midrule
\foreignlanguage{hindi}{महत्व} 
& \foreignlanguage{hindi}{महत्त्व, महत्वम्} \\

\foreignlanguage{hindi}{शहरीकरण} 
& \foreignlanguage{hindi}{शहरीकरण, शहरीकरणम्} \\

\foreignlanguage{hindi}{विकास} 
& \foreignlanguage{hindi}{विकास, विकासः} \\

\foreignlanguage{hindi}{परिवर्तन} 
& \foreignlanguage{hindi}{परिवर्तन, परिवर्तन्} \\
\midrule
\multicolumn{2}{c}{\textbf{Tamil}} \\
\midrule
\TamilText{புவியியல்}
& \TamilText{பூவியியல், புவியலியல்} \\

\TamilText{நிலத்தட்டு}
& \TamilText{நிலத்தட்பு, நிலத் தட்டு} \\

\TamilText{நிலநடுக்கம்}
& \TamilText{நிலநடுக்கம், நிலநடுக்கத்} \\

\TamilText{அதிர்ச்சி}
& \TamilText{அதிர்ச்சி, அதிர்ச்சி\hspace{0.1em}யி} \\
\bottomrule
\end{tabular}

\caption{Illustrative examples from the alternative-spelling dictionary used for Hindi,Telugu and Tamil. Canonical forms are mapped to commonly observed orthographic variants.}
\label{tab:alt_spelling_dict_examples}
\end{table}

Taking a given input text, the attack searches the text for words present in a predefined dictionary (sample entries are shown in Table~\ref{tab:alt_spelling_dict_examples}) using script-aware whole-word matching to prevent partial replacements within longer strings. All the matched words are then substituted with any one of their alternative spellings as per a predefined replacement probability. The process does not modify sentence structure, word sequence, and semantics, but disturbs surface-level orthographic cues and sub word tokenization patterns, and is a specific type of stress test on detectors that are sensitive to fragile spelling regularities.

Formally, let $D$ denote an alternative-spelling dictionary that maps a canonical word $w \in \mathcal{V}$ to a finite set of valid variants
\[
D(w) = \{v_{w,1}, v_{w,2}, \dots, v_{w,m_w}\}.
\]
Let a text sample be represented as a sequence of tokens
\[
\mathbf{x} = (t_1, t_2, \dots, t_n).
\]
We define the set of eligible token positions as
\[
\mathcal{I} = \{\, i \in \{1,\dots,n\} \mid t_i \in \mathcal{V} \,\}.
\]
For each position $i \in \mathcal{I}$, we sample a Bernoulli random variable
\[
z_i \sim \mathrm{Bernoulli}(p),
\]
and apply the replacement rule
\[
t_i' =
\begin{cases}
\mathrm{Unif}(D(t_i)) & \text{if } z_i = 1,\\
t_i & \text{otherwise}.
\end{cases}
\]
Here, $\mathrm{Unif}(D(t_i))$ denotes a uniformly sampled spelling variant from the dictionary entry corresponding to $t_i$. The resulting alternative-spelling attacked text is denoted as
\[
\mathbf{x}^{\text{alt}} = (t_1', t_2', \dots, t_n').
\]

In our implementation, replacements are applied with probability $p = 1.0$ and variants are selected at random, ensuring maximal orthographic diversity while maintaining linguistic validity. Qualitative examples of the alternative spelling attack applied to Telugu, Hindi, and Tamil texts are provided in Tables~\ref{tab:alt_spelling_telugu}, \ref{tab:alt_spelling_hindi}, and \ref{tab:alt_spelling_tamil}.

\subsubsection{Misspelling Attacks}\label{sssec:misspelling}
The misspelling attacks are intended to add natural typographical noise to Hindi, Telugu, and Tamil words by substituting words that are spelled correctly with the most frequently used misspellings without altering the original meaning. The approach employed in this attack is based on a curated misspelling dictionary, in which, for every canonical word, a set of commonly occurring misspellings in natural written language is assigned; sample words from the misspelling dictionary for all three languages are illustrated in Table~\ref{tab:misspelling_dict_examples}. Given a text to work with, we use script-conscious whole-word matching (using Devanagari, Telugu, and Tamil scripts independently) to recognize tokens in the dictionary, and then replace each matched token at random with a misspelled token based on a replacement probability $p$. This process simulates natural human mistakes (e.g., spelling errors and orthographic mistakes) and actively disturbs surface features and tokenization statistics, giving a specific strength test to those detectors sensitive to orthographic regularity.
\begin{table}[t]
\centering
\small
\setlength{\tabcolsep}{6pt}
\renewcommand{\arraystretch}{1.2}

\begin{tabular}{l|l}
\toprule
\textbf{Canonical Form} & \textbf{Misspelling Variants} \\
\midrule
\multicolumn{2}{c}{\textbf{Telugu}} \\
\midrule
\foreignlanguage{telugu}{పరిణామం} & \foreignlanguage{telugu}{పరినామం, పరిణామ} \\
\foreignlanguage{telugu}{సాంకేతికత} & \foreignlanguage{telugu}{సాంకెతికత, సాంకేతికతా} \\
\foreignlanguage{telugu}{అధ్యయనం} & \foreignlanguage{telugu}{అధయనం, అధ్యయన} \\
\foreignlanguage{telugu}{భాషా} & \foreignlanguage{telugu}{భాష, భాషా} \\
\midrule
\multicolumn{2}{c}{\textbf{Hindi}} \\
\midrule
\foreignlanguage{hindi}{महत्वपूर्ण} & \foreignlanguage{hindi}{महत्वपुर्ण, महत्त्वपूर्ण} \\
\foreignlanguage{hindi}{शहरीकरण} & \foreignlanguage{hindi}{शहरिकरण, शहरीकरण} \\
\foreignlanguage{hindi}{पर्यावरणीय} & \foreignlanguage{hindi}{पर्यावर्णीय, पर्यावरणी} \\
\foreignlanguage{hindi}{विश्लेषण} & \foreignlanguage{hindi}{विशलेषण, विश्लेषन} \\
\midrule
\multicolumn{2}{c}{\textbf{Tamil}} \\
\midrule
\TamilText{புவியியல்} & \TamilText{பூவியியல், புவியலியல்} \\
\TamilText{நிலத்தட்டு} & \TamilText{நிலத்தட்பு, நிலத் தட்டு} \\
\TamilText{நிலநடுக்கம்} & \TamilText{நிலநடுக்கம், நிலநடுக்கத்} \\
\TamilText{அதிர்ச்சி} & \TamilText{அதிர்ச்சி, அதிர்ச்சி} \\
\bottomrule
\end{tabular}

\caption{Illustrative examples from the misspelling dictionary used for Hindi, Telugu, and Tamil. Each canonical form is associated with commonly observed misspellings used to simulate natural typographical errors.}
\label{tab:misspelling_dict_examples}
\end{table}

Formally, let $D_{\text{hi}}$, $D_{\text{te}}$, and $D_{\text{ta}}$ denote misspelling dictionaries for Hindi, Telugu, and Tamil, respectively, where each canonical word $w$ maps to a set of misspellings:
\[
D_{\ell}(w) = \{v_{w,1}, \dots, v_{w,m_w}\}, \quad \ell \in \{\text{hi}, \text{te}, \text{ta}\}.
\]
Let a text sample be represented as a token sequence $\mathbf{x}=(t_1,t_2,\dots,t_n)$, and define the eligible positions per language
\[
\mathcal{I}_{\ell}=\{\, i \in \{1,\dots,n\} \mid t_i \in \mathrm{dom}(D_{\ell}) \,\}.
\]
For each $i \in \mathcal{I}_{\ell}$, we sample a Bernoulli random variable $z_i \sim \mathrm{Bernoulli}(p)$ and apply the replacement rule
\[
t_i' =
\begin{cases}
\mathrm{Unif}(D_{\ell}(t_i)) & \text{if } z_i = 1,\\
t_i & \text{otherwise}.
\end{cases}
\]
The resulting misspelling-attacked text is denoted as $\mathbf{x}^{\mathrm{ms}}=(t_1',\dots,t_n')$. In our implementation, we set $p=1.0$ with random variant selection and enforce script-aware whole-word matching to avoid partial substitutions inside longer Indic strings. Representative qualitative examples illustrating the misspelling attack for Telugu, Hindi, and Tamil are provided in Tables~\ref{tab:misspelling_telugu}, \ref{tab:misspelling_hindi}, and \ref{tab:misspelling_tamil}.

\subsubsection{Synonym Swap Attacks}\label{sssec:synonymswap}
Synonym-Swap attacks aim to introduce meaning-preserving lexical variability by substituting some words with semantically identical ones. This attack uses a hand-curated list of Hindi, Telugu, and Tamil synonyms, in which each word is defined as a list of interchangeable word forms; example words for all three languages appear in Table~\ref{tab:synonym_dict_examples}. To ensure we cover as much as possible, we expand the resource to a two-way mapping, so that any synonym cluster can be interchanged with another form. Given a text as input, we perform script-sensitive whole-word matching to find dictionary-covered tokens and, with probability $p$, replace each matched token with a randomly sampled synonym. Because the replacements maintain the structure of the sentence and the intended meaning, but change surface lexical cues and subword tokenization patterns, this attack provides a natural stress test for detectors that rely on brittle word-choice statistics.

\begin{table}[t]
\centering
\small
\setlength{\tabcolsep}{6pt}
\renewcommand{\arraystretch}{1.2}
\resizebox{0.48\textwidth}{!}{%
\begin{tabular}{l|l}
\toprule
\textbf{Token} & \textbf{Synonym Candidates} \\
\midrule
\multicolumn{2}{c}{\textbf{Telugu}} \\
\midrule
\foreignlanguage{telugu}{అవసరం} & \foreignlanguage{telugu}{కావాలి, అవసరమైంది, అవసరకరం} \\
\foreignlanguage{telugu}{ముఖ్యమైన} & \foreignlanguage{telugu}{ప్రాధాన్యమైన, అవసరమైన, ముఖ్యమైనది} \\
\foreignlanguage{telugu}{పరిశోధన} & \foreignlanguage{telugu}{అన్వేషణ, అధ్యయనం, పరిశీలన} \\
\foreignlanguage{telugu}{విశ్లేషణ} & \foreignlanguage{telugu}{పరిశీలన, విశ్లేషణము, అంచనా} \\
\midrule
\multicolumn{2}{c}{\textbf{Hindi}} \\
\midrule
\foreignlanguage{hindi}{महत्वपूर्ण} & \foreignlanguage{hindi}{आवश्यक, प्रमुख, सार्थक} \\
\foreignlanguage{hindi}{विश्लेषण} & \foreignlanguage{hindi}{परीक्षण, समीक्षा, विवेचन} \\
\foreignlanguage{hindi}{विकास} & \foreignlanguage{hindi}{प्रगति, उन्नति, विस्तार} \\
\foreignlanguage{hindi}{परिवर्तन} & \foreignlanguage{hindi}{बदलाव, रूपांतरण, संशोधन} \\
\midrule
\multicolumn{2}{c}{\textbf{Tamil}} \\
\midrule
\TamilText{முக்கியம்} & \TamilText{அவசியம், பிரதானம், முக்கியமானது} \\
\TamilText{ஆய்வு} & \TamilText{பரசோதனை, விசாரணை, ஆராய்ச்சி} \\
\TamilText{வளர்ச்சி} & \TamilText{முன்னேற்றம், மேம்பாடு, விரிவு} \\
\TamilText{மாற்றம்} & \TamilText{திருத்தம், மாற்றம் செய்தல், உருவமாற்றம்} \\
\bottomrule
\end{tabular}
}
\caption{Illustrative examples from the synonym dictionary used for Hindi, Telugu, and Tamil. Each token is associated with a set of meaning-preserving synonym candidates (bidirectional expansion enables swapping among all forms in a cluster).}
\label{tab:synonym_dict_examples}
\end{table}
Formally, let $S_{\text{hi}}$, $S_{\text{te}}$, and $S_{\text{ta}}$ denote synonym resources for Hindi, Telugu, and Tamil, respectively. After bidirectional expansion, we obtain maps $D_{\ell}$ that assign to each form $w$ a set of swappable synonyms:
\[
D_{\ell}(w) \subseteq \mathcal{V}_{\ell}\setminus\{w\}, \quad \ell \in \{\text{hi},\text{te},\text{ta}\}.
\]
Let a text sample be represented as a token sequence $\mathbf{x}=(t_1,t_2,\dots,t_n)$ and define the eligible positions
\[
\mathcal{I}_{\ell}=\{\, i \in \{1,\dots,n\}\mid t_i \in \mathrm{dom}(D_{\ell}) \,\}.
\]
For each $i \in \mathcal{I}_{\ell}$, we sample a Bernoulli variable $z_i \sim \mathrm{Bernoulli}(p)$ and apply the synonym replacement rule
\[
t_i' =
\begin{cases}
\mathrm{Unif}(D_{\ell}(t_i)) & \text{if } z_i = 1,\\
t_i & \text{otherwise},
\end{cases}
\]
where $\mathrm{Unif}(D_{\ell}(t_i))$ denotes a uniformly sampled synonym from the candidate set. The synonym-swapped text is denoted as $\mathbf{x}^{\mathrm{syn}}=(t_1',\dots,t_n')$. In our implementation, we set $p=1.0$ with random selection and enforce Indic script-aware whole-word boundaries to avoid partial replacements within longer strings. Representative qualitative examples illustrating the synonym swap attack for Telugu, Hindi, and Tamil are provided in Tables~\ref{tab:synonym_swap_telugu}, \ref{tab:synonym_swap_hindi}, and \ref{tab:synonym_swap_tamil}.

\subsection{Additional Results}

The detailed results for each evaluation setting defined in the
\textbf{\textit{IndicDetect}} benchmark are presented in this section.
The six settings include In-Distribution, In-Distribution Domain,
In-Distribution Generator, Multi-Domain, Multi-Generator, and Multi-Attack.
Together, these settings examine matched-condition performance, domain
sensitivity, generator sensitivity, cross-domain generalization,
cross-generator generalization, and robustness to adversarial perturbations,
providing a comprehensive view of each detector's strengths and weaknesses.

Results include analysis across Hindi, Telugu, and Tamil, showing similar tendencies across the three languages. 
Neural detectors that are based on XLM-RoBERTa-Base \citep{radford2019gpt2detector} and XLM-RoBERTa-Large \citep{radford2019gpt2detector} remain the top performers, whereas traditional detectors like Fast-DetectGPT \citep{bao2024fastdetectgpt}, Log-Likelihood    \citep{DBLP:journals/corr/abs-1908-09203}, and Log-Rank   \citep{su2023detectllm} perform well on clean data but see their performance decline when subjected to conditions such as cross-generator shift or adversarial attacks. 
The additional results presented here emphasize the capability of the benchmark to reveal different levels of performance depending on the stress factors, thus providing a more comprehensive view of the robustness of detectors in real-world situations.

\clearpage
\begin{figure*}[t]
  \centering
  \includegraphics[width=\textwidth]{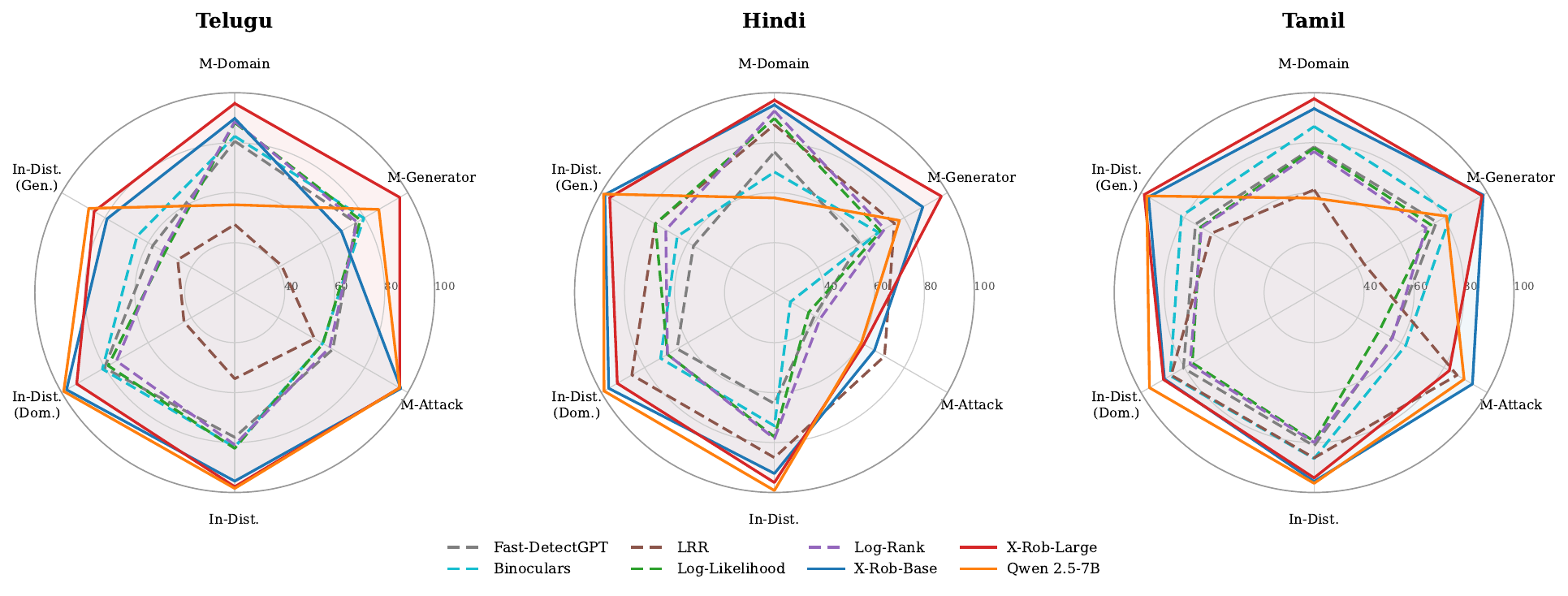}
 \caption{Per-language Macro-F1 of all eight detectors across the six
evaluation settings. Training-free detectors are dashed, and the supervised
and neural detectors (X-Rob-Base, X-Rob-Large, Qwen~2.5-7B) are solid.}
  \label{fig:radar}
\end{figure*}

\newpage

\begin{table*}[t]
\centering
\small
\setlength{\tabcolsep}{10pt}
\renewcommand{\arraystretch}{1.15}

\begin{tabular}{l c}
\toprule
\textbf{Hyperparameter} & \textbf{Value} \\
\midrule
Languages                    & Telugu, Hindi, Tamil \\
Backbone Models              & XLM-RoBERTa-base, XLM-RoBERTa-large \\
Epochs                       & 3 \\
Learning Rate                & $1 \times 10^{-6}$ \\
Batch Size                   & 8 \\
Optimizer                    & AdamW \\
LR Scheduler                 & Constant (no decay) \\
Max Sequence Length          & 512 tokens \\
Loss Function                & Cross-Entropy \\
Random Seed                  & 2023 \\
Validation Split             & 20 examples per class \\
Save Strategy                & Per epoch \\
Early Stopping Patience      & 10 \\
Early Stopping Metric        & Eval Loss (lower is better) \\
Threshold Selection          & Youden's J (max TPR $-$ FPR) \\
\bottomrule
\end{tabular}
\caption{Training hyperparameters shared across Telugu, Hindi, and Tamil models using XLM-RoBERTa-base and XLM-RoBERTa-large variants.}
\label{tab:training_hyperparameters}

\vspace{2em} 

\begin{tabular}{l c}
\toprule
\textbf{Hyperparameter} & \textbf{Value} \\
\midrule
Languages               & Telugu, Hindi, Tamil \\
Backbone Model          & Qwen/Qwen2.5-7B \\
Epochs                  & 1 \\
Learning Rate           & $5 \times 10^{-5}$ \\
Batch Size              & 1 \\
Gradient Accumulation Steps & 16 \\
Effective Batch Size    & 16 \\
Optimizer               & Paged AdamW (8-bit) \\
LR Scheduler            & Cosine decay with warmup \\
Warmup Steps            & 50 \\
Weight Decay            & 0.05 \\
Max Sequence Length     & 512 tokens \\
Quantization            & 4-bit NF4 (QLoRA, double quant) \\
Compute Dtype           & bfloat16 \\
LoRA Rank ($r$)         & 8 \\
LoRA Alpha              & 16 \\
LoRA Dropout            & 0.1 \\
LoRA Target Modules     & \texttt{q\_proj}, \texttt{v\_proj} \\
Early Stopping Patience & 3 \\
Best Model Metric       & Eval Loss (lower is better) \\
\bottomrule
\end{tabular}
\caption{Training hyperparameters for the Qwen~2.5-7B model used across Telugu, Hindi, and Tamil.}
\label{tab:qwen_hyperparameters}
\end{table*}
\clearpage
\begin{table*}[t]
\centering
\small
\renewcommand{\arraystretch}{1.3} 
\addtolength{\tabcolsep}{-2pt}    
\resizebox{\textwidth}{!}{
\begin{tabular}{cccccc}
\toprule
\textbf{Language} & \textbf{Split} & \textbf{Source} & \textbf{TTR} $\uparrow$ & \textbf{Perplexity} $\downarrow$ & \textbf{Top POS (Category 1 / 2 / 3)} \\
\midrule
\multirow{4}{*}{Hindi} & \multirow{2}{*}{Train} & Human & 0.563 & 4.06 & 28.5\% (NOUN) / 18.8\% (ADP) / 11.2\% (ADJ) \\
& & LLM & 0.620 & 2.78 & 29.9\% (NOUN) / 16.7\% (ADP) / 13.9\% (ADJ) \\
\cmidrule{2-6}
& \multirow{2}{*}{Test} & Human & 0.568 & 3.14 & 28.5\% (NOUN) / 17.2\% (ADP) / 11.6\% (ADJ) \\
& & LLM & 0.622 & 2.79 & 29.8\% (NOUN) / 18.0\% (ADP) / 14.0\% (ADJ) \\
\midrule
\multirow{4}{*}{Telugu} & \multirow{2}{*}{Train} & Human & 0.874 & 6.03 & 40.3\% (NOUN) / 17.1\% (VERB) / 15.6\% (PUNCT) \\
& & LLM & 0.891 & 6.88 & 45.8\% (NOUN) / 16.4\% (VERB) / 15.7\% (PUNCT) \\
\cmidrule{2-6}
& \multirow{2}{*}{Test} & Human & 0.877 & 5.03 & 38.0\% (NOUN) / 17.5\% (VERB) / 15.6\% (PUNCT) \\
& & LLM & 0.824 & 6.62 & 45.4\% (NOUN) / 15.3\% (VERB) / 14.1\% (PUNCT) \\
\midrule
\multirow{4}{*}{Tamil} & \multirow{2}{*}{Train} & Human & 0.827 & 5.73 & 37.4\% (NOUN) / 13.1\% (PUNCT) / 10.8\% (PROPN) \\
& & LLM & 0.845 & 6.31 & 38.7\% (NOUN) / 14.6\% (PUNCT) / 10.2\% (VERB) \\
\cmidrule{2-6}
& \multirow{2}{*}{Test} & Human & 0.830 & 5.56 & 34.2\% (NOUN) / 13.1\% (PUNCT) / 11.6\% (PROPN) \\
& & LLM & 0.845 & 6.32 & 39.7\% (NOUN) / 13.5\% (PUNCT) / 10.0\% (ADJ) \\
\bottomrule
\end{tabular}%
} 
\caption{Comprehensive linguistic and statistical baseline across Hindi, Telugu, and Tamil (In-Distribution). TTR: Type-Token Ratio , POS percentages reflect the density of the top three syntactic categories.}
\label{tab:full_indic_baseline}
\end{table*}
\clearpage

\begin{table*}[t]
\centering
\small
\setlength{\tabcolsep}{10pt}
\renewcommand{\arraystretch}{1.15}

\begin{tabular}{l c c}
\toprule
\textbf{Detector} & \textbf{Reference Model} & \textbf{Scoring Model} \\
\midrule
Log-Likelihood & Not Applicable & Qwen-3-4B \\
Log-Rank       & Not Applicable & Qwen-3-4B \\
LRR            & Not Applicable & Qwen-3-4B \\
Fast-DetectGPT  & Qwen-3-4B & Qwen-3-4B \\
Binoculars     & Qwen-3-4B & Qwen-3-4B-Instruct \\
\bottomrule
\end{tabular}

\caption{Source and scoring language models used for training-free detectors. All listed detectors employ Qwen-3-4B as the reference model, with Binoculars using the corresponding instruction-tuned variant for scoring.}
\label{tab:source_reference_detectors}
\end{table*}

\clearpage

\renewcommand{\best}[1]{\textbf{\underline{#1}}}
\renewcommand{\secbest}[1]{\textbf{\textit{#1}}}



\begin{table*}[t]
\centering
\small
\setlength{\tabcolsep}{6pt}
\renewcommand{\arraystretch}{1.1}
\resizebox{0.85\textwidth}{!}{%
\begin{tabular}{lcccccccccc}
\toprule
\multicolumn{11}{c}{\textbf{Leaderboard: LLM-Generated Text Detector Across Domains (Telugu)}} \\
\midrule
\textbf{Detector} &
\multicolumn{2}{c}{\textbf{Academic}} &
\multicolumn{2}{c}{\textbf{News}} &
\multicolumn{2}{c}{\textbf{Creative}} &
\multicolumn{2}{c}{\textbf{Movie}} &
\multicolumn{2}{c}{\textbf{Avg.}} \\
\cmidrule(lr){2-3}\cmidrule(lr){4-5}\cmidrule(lr){6-7}\cmidrule(lr){8-9}\cmidrule(lr){10-11}
& AUROC & $F_1$ & AUROC & $F_1$ & AUROC & $F_1$ & AUROC & $F_1$ & AUROC & $F_1$ \\
\midrule
Fast-DetectGPT & 93.79 & 90.05 & 86.00 & 79.84 & 64.86 & 68.62 & 88.88 & 83.87 & \md{83.38} & \md{80.59} \\
Binoculars     & 92.19 & 89.36 & 87.55 & 78.64 & 81.56 & 80.20 & 85.71 & 82.22 & \hi{86.75} & \md{82.60} \\
LRR            & 53.87 & 68.85 & 46.28 & 83.38 & 84.03 & 36.26 & 41.85 &  0.53 & \lo{56.50} & \lo{47.25} \\
Log-Likelihood & 98.73 & 96.83 & 89.47 & 86.50 & 96.11 & 83.61 & 91.85 & 85.30 & \hi{94.03} & \hi{88.06} \\
Log-Rank       & 98.57 & 96.82 & 88.19 & 83.70 & 96.76 & 85.54 & 92.18 & 87.88 & \hi{93.92} & \hi{88.48} \\
X-Rob-Base     & 99.99 & 89.62 & 99.53 & 80.09 & 100.00 & 92.38 & 99.99 & 96.70 & \hi{\secbest{99.87}} & \hi{\secbest{89.69}} \\
X-Rob-Large    & 100.00 & 98.29 & 99.99 & 87.98 & 100.00 & 99.93 & 99.99 & 96.58 & \hi{\best{99.99}} & \hi{\best{95.69}} \\
Qwen 2.5-7B    & 56.95 & 48.54 & 59.16 & 52.35 & 73.32 & 63.80 & 65.05 & 55.93 & \lo{63.62} & \lo{55.15} \\
\bottomrule
\end{tabular}}
\caption{Per-domain detection on Telugu.
  \colorbox{avgHigh}{\strut Green}~$\geq85$,
  \colorbox{avgMid}{\strut Yellow}~$70$--$84$,
  \colorbox{avgLow}{\strut Red}~$<70$.
  \textbf{\underline{Bold+underline}}~=~1st; \textbf{\textit{bold+italic}}~=~2nd best Avg per column.}
\label{tab:telugu_domains}
\end{table*}

\begin{table*}[t]
\centering
\small
\setlength{\tabcolsep}{6pt}
\renewcommand{\arraystretch}{1.1}
\resizebox{0.85\textwidth}{!}{%
\begin{tabular}{lcccccccccc}
\toprule
\multicolumn{11}{c}{\textbf{Leaderboard: LLM-Generated Text Detector Across Domains (Hindi)}} \\
\midrule
\textbf{Detector} &
\multicolumn{2}{c}{\textbf{Academic}} &
\multicolumn{2}{c}{\textbf{News}} &
\multicolumn{2}{c}{\textbf{Creative}} &
\multicolumn{2}{c}{\textbf{Movie}} &
\multicolumn{2}{c}{\textbf{Avg.}} \\
\cmidrule(lr){2-3}\cmidrule(lr){4-5}\cmidrule(lr){6-7}\cmidrule(lr){8-9}\cmidrule(lr){10-11}
& AUROC & $F_1$ & AUROC & $F_1$ & AUROC & $F_1$ & AUROC & $F_1$ & AUROC & $F_1$ \\
\midrule
Fast-DetectGPT & 72.95 & 81.18 & 62.37 & 76.74 & 54.02 & 85.71 & 64.22 & 61.60 & \lo{63.39} & \md{76.30} \\
Binoculars     & 74.94 & 73.27 & 74.33 & 69.68 & 86.76 & 81.47 & 67.88 & 48.69 & \md{75.97} & \lo{68.27} \\
LRR            & 82.03 & 79.83 & 68.04 & 83.91 & 99.28 & 97.39 & 86.65 & 87.55 & \md{84.00} & \hi{87.17} \\
Log-Likelihood & 95.33 & 94.45 & 82.62 & 82.93 & 93.00 & 90.70 & 83.96 & 90.75 & \hi{88.72} & \hi{89.70} \\
Log-Rank       & 95.76 & 94.63 & 83.04 & 88.00 & 97.61 & 95.01 & 88.50 & 93.27 & \hi{91.22} & \hi{92.72} \\
X-Rob-Base     & 99.80 & 97.65 & 99.98 & 98.85 & 100.00 & 93.02 & 100.00 & 91.01 & \hi{\secbest{99.94}} & \hi{\secbest{95.13}} \\
X-Rob-Large    & 100.00 & 99.93 & 100.00 & 89.62 & 99.99 & 99.27 & 100.00 & 99.33 & \hi{\best{99.99}} & \hi{\best{97.03}} \\
Qwen 2.5-7B    & 54.85 & 42.56 & 55.48 & 52.66 & 81.72 & 72.67 & 75.70 & 63.75 & \lo{66.93} & \lo{57.90} \\
\bottomrule
\end{tabular}}
\caption{Per-domain detection on Hindi.}
\label{tab:hindi_domains}
\end{table*}

\begin{table*}[t]
\centering
\small
\setlength{\tabcolsep}{6pt}
\renewcommand{\arraystretch}{1.1}
\resizebox{0.85\textwidth}{!}{%
\begin{tabular}{lcccccccccc}
\toprule
\multicolumn{11}{c}{\textbf{Leaderboard: LLM-Generated Text Detector Across Domains (Tamil)}} \\
\midrule
\textbf{Detector} &
\multicolumn{2}{c}{\textbf{Academic}} &
\multicolumn{2}{c}{\textbf{News}} &
\multicolumn{2}{c}{\textbf{Creative}} &
\multicolumn{2}{c}{\textbf{Movie}} &
\multicolumn{2}{c}{\textbf{Avg.}} \\
\cmidrule(lr){2-3}\cmidrule(lr){4-5}\cmidrule(lr){6-7}\cmidrule(lr){8-9}\cmidrule(lr){10-11}
& AUROC & $F_1$ & AUROC & $F_1$ & AUROC & $F_1$ & AUROC & $F_1$ & AUROC & $F_1$ \\
\midrule
Fast-DetectGPT & 82.96 & 81.79 & 78.85 & 67.18 & 78.27 & 69.36 & 97.17 & 94.96 & \md{84.31} & \md{78.32} \\
Binoculars     & 91.32 & 91.35 & 85.18 & 78.99 & 86.58 & 79.95 & 98.10 & 95.73 & \hi{90.29} & \hi{86.50} \\
LRR            & 44.84 & 87.55 & 39.24 & 84.86 & 44.28 &  3.27 & 54.29 & 69.04 & \lo{45.66} & \lo{61.18} \\
Log-Likelihood & 88.32 & 88.98 & 69.76 & 70.34 & 70.45 & 52.75 & 99.84 & 99.23 & \md{82.09} & \md{77.82} \\
Log-Rank       & 85.30 & 87.67 & 66.97 & 68.22 & 68.59 & 51.20 & 99.80 & 98.72 & \md{80.16} & \md{76.45} \\
X-Rob-Base     & 99.99 & 89.60 & 99.87 & 92.65 & 100.00 & 93.49 & 100.00 & 98.68 & \hi{\secbest{99.96}} & \hi{\secbest{93.60}} \\
X-Rob-Large    & 99.99 & 99.25 & 100.00 & 96.21 & 100.00 & 97.40 & 100.00 & 97.52 & \hi{\best{99.99}} & \hi{\best{97.59}} \\
Qwen 2.5-7B    & 49.43 & 49.19 & 52.59 & 51.43 & 86.74 & 75.44 & 62.84 & 55.20 & \lo{62.90} & \lo{57.81} \\
\bottomrule
\end{tabular}}
\caption{Per-domain detection on Tamil.}
\label{tab:tamil_domains}
\end{table*}

\clearpage


\begin{table*}[t]
\centering
\small
\setlength{\tabcolsep}{6pt}
\renewcommand{\arraystretch}{1.1}
\resizebox{0.85\textwidth}{!}{%
\begin{tabular}{lcccccccc}
\toprule
\multicolumn{9}{c}{\textbf{Leaderboard: LLM-Generated Text Detection by Source Model (Telugu)}} \\
\midrule
\textbf{Detector} &
\multicolumn{2}{c}{\textbf{GPT-4.1}} &
\multicolumn{2}{c}{\textbf{Qwen-Plus}} &
\multicolumn{2}{c}{\textbf{DeepSeek-v3}} &
\multicolumn{2}{c}{\textbf{Avg.}} \\
\cmidrule(lr){2-3}\cmidrule(lr){4-5}\cmidrule(lr){6-7}\cmidrule(lr){8-9}
& AUROC & $F_1$ & AUROC & $F_1$ & AUROC & $F_1$ & AUROC & $F_1$ \\
\midrule
Fast-DetectGPT & 69.51 & 60.92 & 97.42 & 91.60 & 83.23 & 75.52 & \md{83.38} & \md{76.01} \\
Binoculars     & 73.01 & 66.98 & 97.92 & 92.49 & 87.53 & 79.55 & \hi{86.15} & \md{79.67} \\
LRR            & 65.33 & 61.98 & 43.58 &  0.00 & 56.85 & 63.02 & \lo{55.25} & \lo{41.66} \\
Log-Likelihood & 73.56 & 67.47 & 95.14 & 88.56 & 85.97 & 78.28 & \md{84.89} & \md{78.10} \\
Log-Rank       & 73.87 & 69.13 & 92.05 & 84.45 & 84.74 & 76.03 & \md{83.55} & \md{76.53} \\
X-Rob-Base     & 42.68 &  9.31 & 99.90 & 98.62 & 99.95 & 99.82 & \md{80.84} & \lo{69.25} \\
X-Rob-Large    & 99.99 & 97.97 & 99.99 & 95.84 & 100.00 & 94.96 & \hi{\best{99.99}} & \hi{\best{96.25}} \\
Qwen 2.5-7B    & 86.73 & 78.42 & 99.27 & 96.42 & 93.05 & 84.95 & \hi{\secbest{93.01}} & \hi{\secbest{86.59}} \\
\bottomrule
\end{tabular}}
\caption{Per-generator detection on Telugu.}
\label{tab:telugu_llms}
\end{table*}

\begin{table*}[t]
\centering
\small
\setlength{\tabcolsep}{6pt}
\renewcommand{\arraystretch}{1.1}
\resizebox{0.85\textwidth}{!}{%
\begin{tabular}{lcccccccc}
\toprule
\multicolumn{9}{c}{\textbf{Leaderboard: LLM-Generated Text Detection by Source Model (Hindi)}} \\
\midrule
\textbf{Detector} &
\multicolumn{2}{c}{\textbf{GPT-4.1}} &
\multicolumn{2}{c}{\textbf{Qwen-Plus}} &
\multicolumn{2}{c}{\textbf{DeepSeek-v3}} &
\multicolumn{2}{c}{\textbf{Avg.}} \\
\cmidrule(lr){2-3}\cmidrule(lr){4-5}\cmidrule(lr){6-7}\cmidrule(lr){8-9}
& AUROC & $F_1$ & AUROC & $F_1$ & AUROC & $F_1$ & AUROC & $F_1$ \\
\midrule
Fast-DetectGPT & 53.99 & 49.78 & 80.41 & 74.68 & 51.12 & 53.61 & \lo{61.83} & \lo{59.35} \\
Binoculars     & 67.23 & 59.80 & 93.12 & 85.51 & 67.73 & 59.09 & \md{76.02} & \lo{68.13} \\
LRR            & 78.81 & 75.02 & 70.94 & 71.74 & 83.21 & 79.62 & \md{77.65} & \md{75.45} \\
Log-Likelihood & 69.24 & 66.10 & 84.87 & 74.58 & 72.60 & 66.68 & \md{75.56} & \lo{69.12} \\
Log-Rank       & 71.70 & 68.48 & 83.54 & 75.99 & 75.64 & 68.43 & \md{76.95} & \md{70.96} \\
X-Rob-Base     & 37.46 & 66.40 & 99.99 & 99.90 & 99.74 & 99.42 & \md{79.06} & \hi{\secbest{88.57}} \\
X-Rob-Large    & 99.99 & 95.86 & 99.99 & 97.97 & 100.00 & 97.62 & \hi{\best{99.99}} & \hi{\best{97.15}} \\
Qwen 2.5-7B    & 81.28 & 73.52 & 91.05 & 82.36 & 84.70 & 77.47 & \hi{\secbest{85.67}} & \md{77.78} \\
\bottomrule
\end{tabular}}
\caption{Per-generator detection on Hindi.}
\label{tab:hindi_llms}
\end{table*}

\begin{table*}[t]
\centering
\small
\setlength{\tabcolsep}{6pt}
\renewcommand{\arraystretch}{1.1}
\resizebox{0.85\textwidth}{!}{%
\begin{tabular}{lcccccccc}
\toprule
\multicolumn{9}{c}{\textbf{Leaderboard: LLM-Generated Text Detection by Source Model (Tamil)}} \\
\midrule
\textbf{Detector} &
\multicolumn{2}{c}{\textbf{GPT-4.1}} &
\multicolumn{2}{c}{\textbf{Qwen-Plus}} &
\multicolumn{2}{c}{\textbf{DeepSeek-v3}} &
\multicolumn{2}{c}{\textbf{Avg.}} \\
\cmidrule(lr){2-3}\cmidrule(lr){4-5}\cmidrule(lr){6-7}\cmidrule(lr){8-9}
& AUROC & $F_1$ & AUROC & $F_1$ & AUROC & $F_1$ & AUROC & $F_1$ \\
\midrule
Fast-DetectGPT & 71.87 & 62.77 & 97.28 & 91.88 & 83.11 & 75.11 & \md{84.08} & \md{76.58} \\
Binoculars     & 80.82 & 72.37 & 98.77 & 94.62 & 90.23 & 82.77 & \hi{89.94} & \md{83.25} \\
LRR            & 41.80 &  0.00 & 43.32 & 67.55 & 56.03 & 61.60 & \lo{47.05} & \lo{43.05} \\
Log-Likelihood & 63.39 & 56.79 & 93.35 & 89.36 & 84.31 & 76.13 & \md{80.35} & \md{74.09} \\
Log-Rank       & 61.06 & 54.06 & 90.20 & 86.76 & 82.47 & 74.41 & \md{77.91} & \md{71.74} \\
X-Rob-Base     & 99.84 & 96.50 & 99.98 & 97.80 & 99.89 & 98.52 & \hi{\secbest{99.90}} & \hi{\secbest{97.60}} \\
X-Rob-Large    & 99.99 & 98.64 & 100.00 & 99.51 & 100.00 & 94.76 & \hi{\best{99.99}} & \hi{\best{97.63}} \\
Qwen 2.5-7B    & 85.25 & 78.15 & 91.95 & 84.65 & 88.08 & 80.87 & \hi{88.42} & \md{81.22} \\
\bottomrule
\end{tabular}}
\caption{Per-generator detection on Tamil.}
\label{tab:tamil_llms}
\end{table*}

\clearpage


\begin{table*}[t]
\centering
\small
\setlength{\tabcolsep}{4pt}
\renewcommand{\arraystretch}{1.1}
\resizebox{\textwidth}{!}{%
\begin{tabular}{lcccccccccccccccc}
\toprule
\multicolumn{17}{c}{\textbf{Leaderboard: LLM-Generated Text Detector Under Adversarial Attacks (Telugu)}} \\
\midrule
\textbf{Detector} &
\multicolumn{2}{c}{\textbf{Paraphrase}} &
\multicolumn{2}{c}{\textbf{Perturbation}} &
\multicolumn{2}{c}{\textbf{White Space}} &
\multicolumn{2}{c}{\textbf{Alt.\ Spelling}} &
\multicolumn{2}{c}{\textbf{Insert Para.}} &
\multicolumn{2}{c}{\textbf{Misspelling}} &
\multicolumn{2}{c}{\textbf{Synonym Swap}} &
\multicolumn{2}{c}{\textbf{Avg.}} \\
\cmidrule(lr){2-3}\cmidrule(lr){4-5}\cmidrule(lr){6-7}
\cmidrule(lr){8-9}\cmidrule(lr){10-11}\cmidrule(lr){12-13}
\cmidrule(lr){14-15}\cmidrule(lr){16-17}
& AUC & $F_1$ & AUC & $F_1$ & AUC & $F_1$
& AUC & $F_1$ & AUC & $F_1$ & AUC & $F_1$
& AUC & $F_1$ & AUC & $F_1$ \\
\midrule
Fast-DetectGPT & 87.50 & 85.69 & 81.95 & 79.85 & 65.28 & 57.45 & 75.56 & 71.16 & 79.25 & 76.74 & 51.99 & 34.64 & 63.94 & 51.67 & \md{72.21} & \lo{65.31} \\
Binoculars     & 89.60 & 85.62 & 84.60 & 79.91 & 56.63 & 40.31 & 78.13 & 68.99 & 81.33 & 75.33 & 47.77 & 23.27 & 68.66 & 51.13 & \md{72.38} & \lo{60.65} \\
LRR            & 48.90 & 35.10 & 54.55 & 46.06 & 63.63 & 59.73 & 66.89 & 63.84 & 59.29 & 53.83 & 84.07 & 86.51 & 59.20 & 53.01 & \lo{62.36} & \lo{56.86} \\
Log-Likelihood & 85.31 & 82.51 & 83.33 & 80.80 & 71.88 & 60.85 & 77.27 & 70.18 & 83.04 & 79.95 & 61.45 & 33.29 & 60.02 & 17.80 & \md{74.61} & \lo{60.76} \\
Log-Rank       & 82.62 & 80.74 & 81.79 & 78.57 & 72.01 & 62.36 & 77.46 & 71.20 & 82.22 & 79.02 & 67.54 & 49.74 & 60.07 & 25.74 & \md{74.81} & \lo{63.91} \\
X-Rob-Base     & 99.98 & 94.20 & 99.93 & 96.45 & 99.99 & 97.51 & 99.99 & 98.16 & 99.99 & 97.51 & 99.98 & 94.29 & 99.99 & 99.25 & \hi{\secbest{99.97}} & \hi{\best{96.76}} \\
X-Rob-Large    & 99.99 & 98.18 & 99.95 & 97.21 & 99.99 & 98.40 & 99.99 & 96.07 & 99.99 & 98.40 & 99.98 & 86.70 & 99.99 & 98.60 & \hi{\best{99.98}} & \hi{96.22} \\
Qwen 2.5-7B    & 99.26 & 95.51 & 99.58 & 97.28 & 99.42 & 96.85 & 99.53 & 97.23 & 99.63 & 98.08 & 98.96 & 93.49 & 99.29 & 95.29 & \hi{99.38} & \hi{\secbest{96.24}} \\
\bottomrule
\end{tabular}}
\caption{Detection on Telugu under seven adversarial attacks.}
\label{tab:telugu_attacks}
\end{table*}

\begin{table*}[t]
\centering
\small
\setlength{\tabcolsep}{4pt}
\renewcommand{\arraystretch}{1.1}
\resizebox{\textwidth}{!}{%
\begin{tabular}{lcccccccccccccccc}
\toprule
\multicolumn{17}{c}{\textbf{Leaderboard: LLM-Generated Text Detector Under Adversarial Attacks (Hindi)}} \\
\midrule
\textbf{Detector} &
\multicolumn{2}{c}{\textbf{Paraphrase}} &
\multicolumn{2}{c}{\textbf{Perturbation}} &
\multicolumn{2}{c}{\textbf{White Space}} &
\multicolumn{2}{c}{\textbf{Alt.\ Spelling}} &
\multicolumn{2}{c}{\textbf{Insert Para.}} &
\multicolumn{2}{c}{\textbf{Misspelling}} &
\multicolumn{2}{c}{\textbf{Synonym Swap}} &
\multicolumn{2}{c}{\textbf{Avg.}} \\
\cmidrule(lr){2-3}\cmidrule(lr){4-5}\cmidrule(lr){6-7}
\cmidrule(lr){8-9}\cmidrule(lr){10-11}\cmidrule(lr){12-13}
\cmidrule(lr){14-15}\cmidrule(lr){16-17}
& AUC & $F_1$ & AUC & $F_1$ & AUC & $F_1$
& AUC & $F_1$ & AUC & $F_1$ & AUC & $F_1$
& AUC & $F_1$ & AUC & $F_1$ \\
\midrule
Fast-DetectGPT & 69.13 & 73.61 & 42.46 & 39.76 & 61.84 & 66.92 &  9.47 &  1.09 & 79.25 & 82.08 & 12.36 &  2.56 & 18.21 &  3.80 & \lo{41.81} & \lo{38.54} \\
Binoculars     & 78.85 & 76.01 & 49.25 & 37.38 & 76.03 & 72.65 &  5.72 &  0.37 &   5.71 &   0.43 & 13.61 &  3.52 & 20.85 &  1.41 & \lo{35.71} & \lo{27.39} \\
LRR            & 73.97 & 84.74 & 60.64 & 69.85 & 77.65 & 86.27 & 75.31 & 83.70 & 73.67 & 83.34 & 74.17 & 81.09 & 26.87 &  7.18 & \lo{66.04} & \md{\best{70.88}} \\
Log-Likelihood & 76.90 & 79.08 & 48.34 & 34.33 &   67.63 &   56.30 & 27.78 &  1.01 & 74.30 & 76.23 & 31.32 &  3.84 & 14.67 &  0.00 & \lo{48.70} & \lo{35.82} \\
Log-Rank       & 77.21 & 78.65 & 50.35 & 34.57 & 76.96 & 78.37 & 36.56 &  7.01 & 75.06 & 76.57 & 39.97 & 12.28 & 16.56 &  0.00 & \lo{53.23} & \lo{41.06} \\
X-Rob-Base     & 99.98 & 92.42 & 99.97 & 92.46 & 99.97 & 91.99 & 95.92 & 30.39 & 99.90 & 94.72 & 84.15 & 20.02 & 98.98 & 41.50 & \hi{\secbest{96.98}} & \lo{\secbest{66.21}} \\
X-Rob-Large    & 99.99 & 94.50 & 99.99 & 81.72 & 99.99 & 95.38 & 99.14 & 20.11 & 99.99 & 95.67 & 98.80 & 21.89 & 97.71 & 20.00 & \hi{\best{99.37}} & \lo{61.32} \\
Qwen 2.5-7B    & 99.91 & 96.10 & 99.64 & 82.75 & 99.96 & 98.93 & 90.84 & 24.44 & 97.23 & 78.41 & 81.03 & 20.21 & 89.39 & 20.42 & \hi{94.00} & \lo{60.18} \\
\bottomrule
\end{tabular}}
\caption{Detection on Hindi under seven adversarial attacks.}
\label{tab:hindi_attacks}
\end{table*}

\begin{table*}[t]
\centering
\small
\setlength{\tabcolsep}{4pt}
\renewcommand{\arraystretch}{1.1}
\resizebox{\textwidth}{!}{%
\begin{tabular}{lcccccccccccccccc}
\toprule
\multicolumn{17}{c}{\textbf{Leaderboard: LLM-Generated Text Detector Under Adversarial Attacks (Tamil)}} \\
\midrule
\textbf{Detector} &
\multicolumn{2}{c}{\textbf{Paraphrase}} &
\multicolumn{2}{c}{\textbf{Perturbation}} &
\multicolumn{2}{c}{\textbf{White Space}} &
\multicolumn{2}{c}{\textbf{Alt.\ Spelling}} &
\multicolumn{2}{c}{\textbf{Insert Para.}} &
\multicolumn{2}{c}{\textbf{Misspelling}} &
\multicolumn{2}{c}{\textbf{Synonym Swap}} &
\multicolumn{2}{c}{\textbf{Avg.}} \\
\cmidrule(lr){2-3}\cmidrule(lr){4-5}\cmidrule(lr){6-7}
\cmidrule(lr){8-9}\cmidrule(lr){10-11}\cmidrule(lr){12-13}
\cmidrule(lr){14-15}\cmidrule(lr){16-17}
& AUC & $F_1$ & AUC & $F_1$ & AUC & $F_1$
& AUC & $F_1$ & AUC & $F_1$ & AUC & $F_1$
& AUC & $F_1$ & AUC & $F_1$ \\
\midrule
Fast-DetectGPT & 84.27 & 81.75 & 78.56 & 75.02 & 74.93 & 70.60 & 76.66 & 71.76 & 76.15 & 72.52 & 13.56 &  1.55 & 42.71 & 18.55 & \lo{63.83} & \lo{55.96} \\
Binoculars     & 90.04 & 87.21 & 85.49 & 82.29 & 81.30 & 77.32 & 83.18 & 79.49 & 83.76 & 80.54 & 14.53 &  2.42 & 50.09 & 27.05 & \lo{69.77} & \lo{62.33} \\
LRR            & 46.01 & 86.03 & 49.93 & 86.09 & 46.75 & 85.95 & 55.28 & 86.14 & 42.05 & 85.81 & 94.25 & 86.22 & 58.33 & 86.19 & \lo{56.08} & \hi{86.06} \\
Log-Likelihood & 80.39 & 79.54 & 71.80 & 69.05 & 67.61 & 60.27 & 71.84 & 66.59 & 75.34 & 74.61 & 24.15 &  0.56 & 35.77 &  0.94 & \lo{60.98} & \lo{50.22} \\
Log-Rank       & 77.87 & 80.30 & 70.15 & 72.31 & 65.74 & 65.35 & 70.83 & 72.58 & 72.41 & 75.88 & 42.59 & 20.29 & 39.11 &  6.98 & \lo{62.67} & \lo{56.24} \\
X-Rob-Base     & 99.98 & 95.22 & 99.98 & 94.73 & 99.97 & 93.94 & 99.97 & 94.70 & 99.98 & 96.37 & 99.91 & 88.10 & 99.95 & 89.12 & \hi{\secbest{99.96}} & \hi{\best{93.16}} \\
X-Rob-Large    & 99.99 & 90.07 & 99.99 & 86.74 & 99.97 & 71.68 & 99.99 & 89.56 & 99.99 & 83.58 & 99.99 & 72.79 & 99.99 & 82.79 & \hi{\best{99.98}} & \md{82.45} \\
Qwen 2.5-7B    & 99.44 & 96.40 & 99.18 & 93.56 & 98.67 & 86.99 & 99.26 & 93.80 & 99.15 & 92.32 & 98.08 & 81.27 & 97.33 & 80.86 & \hi{98.73} & \hi{\secbest{89.31}} \\
\bottomrule
\end{tabular}}
\caption{Detection on Tamil under seven adversarial attacks.}
\label{tab:tamil_attacks}
\end{table*}

\newcolumntype{Y}{>{\RaggedRight\arraybackslash}X}

\newcommand{\PromptField}[1]{\textbf{#1}}
\newcommand{\PromptMono}[1]{\texttt{#1}}

\begin{table*}[t]
\centering
\scriptsize
\setlength{\tabcolsep}{3pt}
\renewcommand{\arraystretch}{1.1}
\begin{tabularx}{\textwidth}{@{}
  p{1.8cm}
  >{\RaggedRight\arraybackslash}X
@{}}
\toprule
\textbf{Domain} & \textbf{Prompt Template} \\
\bottomrule
\end{tabularx}
\caption{Keyword-pipeline prompting templates for Telugu data generation across domains.}
\label{tab:telugu_prompts}
\end{table*}

\begin{table*}[t]
\centering
\scriptsize
\setlength{\tabcolsep}{3pt}
\renewcommand{\arraystretch}{1.1}
\begin{tabularx}{\textwidth}{@{}
  p{1.8cm}
  >{\RaggedRight\arraybackslash}X
@{}}
\toprule
\textbf{Domain} & \textbf{Prompt Template} \\
\bottomrule
\end{tabularx}
\caption{Keyword-pipeline prompting templates for Hindi data generation across domains.}
\label{tab:hindi_prompts}
\end{table*}

\begin{table*}[t]
\centering
\scriptsize
\setlength{\tabcolsep}{3pt}
\renewcommand{\arraystretch}{1.1}
\begin{tabularx}{\textwidth}{@{}
  p{1.8cm}
  >{\RaggedRight\arraybackslash}X
@{}}
\toprule
\textbf{Domain} & \textbf{Prompt Template} \\
\midrule

\textbf{Academic} &
\begin{minipage}[t]{\linewidth}\scriptsize\RaggedRight\sloppy
\textbf{Keyword:}
\texttt{build\_keyword\_prompt}\allowbreak\texttt{(sample\_text)}\\[1pt]
\TamilText{கீழே உள்ள உரைப் பகுதிலிருந்து}
\TamilText{துல்லியமாக 4 அறிவியல் அல்லது}
\TamilText{கல்விசார் முக்கியச் சொற்களை தமிழில் வழங்கவும்.}
\TamilText{விளக்கங்கள் அல்லது பிற வாக்கியங்கள் எழுத வேண்டாம்:}
\texttt{\{sample\_text\}}\\[3pt]
\textbf{Article:}
\texttt{build\_article\_prompt}\allowbreak\texttt{(keywords)}\\[1pt]
\TamilText{இந்த 4 முக்கியச் சொற்களை அடிப்படையாகக் கொண்டு}
\TamilText{முழுவதும் தமிழ் மொழியில்}
\TamilText{ஒரு புதிய, விரிவான கல்விசார் கட்டுரையை மட்டும் எழுதவும்.}
\TamilText{கட்டுரை குறைந்தது 30 வாக்கியங்கள் இருக்க வேண்டும்.}
\TamilText{ஆராய்ச்சிசார் பாணி,}
\TamilText{தர்க்கபூர்வமான அமைப்பு,}
\TamilText{தெளிவான பகுப்பாய்வு இருக்க வேண்டும்.}
\TamilText{அதே சொற்களை மீண்டும் மீண்டும் பயன்படுத்த வேண்டாம்.}
\TamilText{பல்வகை வாக்கிய அமைப்பைப் பயன்படுத்தவும்.}
\TamilText{மேற்கோள்கள், விளக்கங்கள், அல்லது முன்னுரை வழங்க வேண்டாம்.}
\TamilText{கல்விசார் கட்டுரையை மட்டும் எழுதவும்.}
\TamilText{முக்கியச் சொற்கள்:} \texttt{\{keywords\}}
\TamilText{புதிய கல்விசார் கட்டுரை:}\\[3pt]
\textbf{Extend:}
\texttt{build\_extend\_prompt}\allowbreak\texttt{(keywords,}\allowbreak\texttt{ existing\_text)}\\[1pt]
\TamilText{இந்த முக்கியச் சொற்களை அடிப்படையாகக் கொண்டு}
\TamilText{கீழே உள்ள தமிழ் கல்விசார் கட்டுரையை தொடரவும்.}
\TamilText{ஏற்கனவே எழுதப்பட்ட விஷயங்களை மீண்டும் எழுத வேண்டாம்.}
\TamilText{புதிய பகுப்பாய்வு,}
\TamilText{புதிய ஆராய்ச்சி கோணங்கள்,}
\TamilText{புதிய தகவல்கள் மட்டும் சேர்க்கவும்.}
\TamilText{தர்க்கபூர்வமான பாணி தொடர வேண்டும்.}
\TamilText{குறைந்தது 15 புதிய வாக்கியங்கள் எழுதவும்.}
\TamilText{முக்கியச் சொற்கள்:} \texttt{\{keywords\}}
\TamilText{இதுவரை:} \texttt{\{existing\_text\}}
\TamilText{தொடர்ச்சி:}
\end{minipage}\\
\bottomrule
\end{tabularx}
\caption{Keyword-pipeline prompting templates for Tamil data generation across domains.}
\label{tab:tamil_prompts}
\end{table*}

\newpage
\begin{table*}[t]
\centering
\small
\setlength{\tabcolsep}{6pt}
\renewcommand{\arraystretch}{1.2}

\begin{tabular}{>{\RaggedRight\arraybackslash}p{0.97\textwidth}}
\toprule
\multicolumn{1}{c}{\textbf{Telugu Paraphrase Example (Original $\rightarrow$ Chinese $\rightarrow$ Original)}} \\
\midrule

\textbf{Original:} \\
\foreignlanguage{telugu}{
సాహిత్య, విజ్ఞాన రంగాల్లో ముద్రణ ప్రక్రియకు కేంద్రస్థానం ఉంది. ఇది రచనలను విస్తరించడంలో అంశపూరిత పాత్ర పోషిస్తుంది. ప్రాచీన భారతదేశంలో ముఖ్యంగా హస్తప్రతులు ప్రాచుర్యంలో ఉండగా, యూరోపియన్ సాంప్రదాయాలలో ముద్రణ పద్దతులు వచ్చి విజ్ఞాన వ్యాప్తికి మౌలిక మద్దతునిచ్చాయి. 15వ శతాబ్దంలో గుటెన్బర్గ్ ప్రెస్ ఆవిష్కరణ తర్వాత పాఠ్య గ్రంథాలు ఎక్కువ సంఖ్యలో అందుబాటులోకి రాగా, అభివృద్ధి చెందిన దేశాల్లో విద్యారంగమే కాక సామాజిక సంస్కృతి కూడా ప్రగతిశీల మార్పులకు లోను అయింది. ఆధునిక కాలంలో పుస్తకాల ముద్రణ, అనువాద ప్రక్రియ అనుసంధానంతో మరింత విస్తృతమవుతోంది. వచనాలను ఒక భాషలోనుంచి మరో భాషలోకి మార్చడాన్ని సూచించే అనువాదం, జ్ఞానాభివృద్ధికి పెనుమార్గం చూపింది. సామాజిక, సాంస్కృతిక పరస్పర సంబంధాలను బలోపేతం చేయడంలో ఇది విశేష పాత్రను నిర్వర్తించింది. ఉన్నత విద్యాసంస్థల్లో, విశ్వవిద్యాలయ గ్రంథాలయం వంటి వనరుల ద్వారా ప్రయోజనాన్ని ఆనందించవచ్చు. విదేశీ రచనల జ్ఞానం అనువాద ప్రక్రియ కారణంగా సమాజానికి అందుతుంది. విద్యా విధానంలో మిషన్ ముఖ్యమైన ఆవశ్యకం. సాధికారత, విద్యార్థులకు నూతన ఆలోచనల్ని ఇవ్వడం వంటి లక్ష్యాలు విద్యా ప్రణాళికలో భాగంగా మారాయి. కొన్ని విద్యా సంస్థలు విశేష మిషన్ స్టేట్మెంట్లను రూపొందించి వాటిని ఆచరణలో పెట్టిన తీరు పరిశీలనార్హం. విద్యమార్గాన్ని విశాలంగా చేయడంలో, జ్ఞానం పంపిణీలో ఉపకరణాలుగా ముద్రిత గ్రంథాలు ఉపయోగపడతాయి. ఉపయోగకరమైన సమాచారాన్ని సమగ్రంగా అందించడంలో మిషన్ యొక్క ప్రాముఖ్యతను గుర్తించాలి. గ్రంథరచన ఒక సమగ్ర శాస్త్రీయ ప్రక్రియ. రచయితలు పరిశోధన, విమర్శన విశ్లేషణల ద్వారా సమాచారాన్ని సంకలనం చేస్తారు. విజ్ఞాన పరంగా, చిన్న పుస్తకం నుండి విశాల గ్రంధాలయ విలువైన వనరుల వరకు వివిధ స్థాయిలో రచనలు ప్రాచుర్యం పొందాయి. సాంప్రదాయము, ఆధునికత కలబోతతో రూపొందిన గ్రంథాలు సమాజానికి మార్గదర్శిగా నిలుస్తున్నాయి.
}
\\[0.4em]

\textbf{Back-Translated:} \\
\bottomrule
\end{tabular}

\caption{Illustrative Telugu paraphrase example generated via a Chinese pivot (Original $\rightarrow$ Chinese $\rightarrow$ Original) to simulate paraphrasing-based adversarial attacks. Back-translated segments are shown in \textbf{bold} for clarity.}
\label{tab:paraphrase_telugu}
\end{table*}

\begin{table*}[t]
\centering
\small
\setlength{\tabcolsep}{6pt}
\renewcommand{\arraystretch}{1.2}

\begin{tabular}{>{\RaggedRight\arraybackslash}p{0.97\textwidth}}
\toprule
\multicolumn{1}{c}{\textbf{Hindi Paraphrase Example (Original $\rightarrow$ Chinese $\rightarrow$ Original)}} \\
\midrule

\textbf{Original:} \\
\bottomrule
\end{tabular}

\caption{Illustrative Hindi paraphrase example generated via a Chinese pivot (Original $\rightarrow$ Chinese $\rightarrow$ Original) to simulate paraphrasing-based adversarial attacks. Back-translated segments are shown in \textbf{bold} for clarity.}

\label{tab:paraphrase_hindi}
\end{table*}

\begin{table*}[t]
\centering
\small
\setlength{\tabcolsep}{6pt}
\renewcommand{\arraystretch}{1.2}

\begin{tabular}{>{\RaggedRight\arraybackslash}p{0.97\textwidth}}
\toprule

\multicolumn{1}{c}{
\textbf{Tamil Paraphrase Example (Original $\rightarrow$ Chinese $\rightarrow$ Original)}
} \\

\midrule

\textbf{Original:} \\

\TamilText{%
  புவியியல் அறிவியல் துறையில் நிலத்தட்டுகளும், அவற்றைச் சூழவுள்ள இயக்கங்களும் மிக அதிக முக்கியத்துவம் பெறுகின்றன. நிலத்தட்டு என்பது பூமியின் மேற்பரப்பை மூடியிருக்கும் பெரும் பாறைத் தகடுகளாகும். இவை வெவ்வேறு பரப்பளவிலும், தடிமனுமாகவும் காணப்படுகின்றன. பூமியின் மேல்பகுதி மன்றாட் சூழல் போன்ற வெப்ப, அழுத்த வேறுபாடுகளினால், இவை சிறிது சிறிதாக நகர்கின்றன. இந்த நிலத்தட்டுகளுக்கிடையில் இருக்கும் எல்லைகள் வெடிப்புகள், இடர்பாடுகள், மோதல்கள் மற்றும் ஊடாடும் கருக்கைகளால் குறிப்பிட்ட இடங்களில் மிகுந்த அழுத்தம் ஏற்படுகிறது. சில நேரங்களில், நிலத்தட்டுகள் விரிவடையும் பகுதியில் அவை மாறுபாடு கொண்ட பதிப்புகளாக பிரிந்து போகும். இன்னும் சில நேரங்களில் தகடுகள் ஒன்றினொன்று மேல் செல்கின்றன அல்லது பாதியில் மோதுகின்றன. இந்த இயக்கங்கள் நிலைபெறவில்லை என்றால், திடீரென ஆற்றல் வெளிப்படுகிறது. இப்படியான ஆற்றல் வெளியீடு நிலநடுக்கத்தை உருவாக்குகிறது. நிலநடுக்கத்தின் போது பூமியின் மேற்பரப்பு பிசிறி, படிப்படியாக அதிர்ச்சி அலைகள் சுற்றியுள்ள பகுதிகளிலும் பரவுகின்றன. இது கட்டிடங்கள் மற்றும் பிற கட்டமைப்புகளில் சேதம் ஏற்படுத்த முடியும். டெக்டானிக் பிளேட் இயக்கங்கள் காரணமாக பெரும்பாலான நிலநடுக்கங்கள் ஏற்படுகின்றன. பூமியில் இருக்கும் இது போன்ற இயக்கங்கள் பெரும்பாலும் அகன்று ஆழத்தில் அல்லது புவியின் எல்லைகளில் நிகழ்கின்றன. மேலோட்டமான இயக்கங்கள் எழும்பும் பொழுது, சந்திக்கப்பட்ட இடத்தில் பெரும் அழுத்தத் தாக்கம் தெரிகிறது. சில சிக்கலான நிலைமைகளில், திடீரென ஆழமான பாகங்களில் நோடுகள் முறிவு பெறும்படி செய்கின்றன. இவ்வாறு நுழைந்தும் வெளியேறும் ஆற்றல் பூமியின் மேற்பரப்பை அதிர வைக்கும். சில நிலநடுக்கங்கள் கடலுக்கடியில் அல்லது குறுகிய ஆழத்தில் நிகழ்கின்றன. இவை கடல் நீரில் பெரும் அலைகள் அல்லது அதிர்ச்சியை உண்டாக்கும்.
  }\\

\vspace{0.4em}
\textbf{Back-Translated:} \\
\bottomrule
\end{tabular}

\caption{
Illustrative Tamil paraphrase example generated via a Chinese pivot
(Original $\rightarrow$ Chinese $\rightarrow$ Original) to simulate
paraphrasing-based adversarial attacks. Back-translated segments are shown
in \textbf{bold} for clarity.
}

\label{tab:paraphrase_tamil}

\end{table*}

\begin{table*}[t]
\centering
\small
\setlength{\tabcolsep}{6pt}
\renewcommand{\arraystretch}{1.2}

\begin{tabular}{>{\RaggedRight\arraybackslash}p{0.97\textwidth}}
\toprule
\multicolumn{1}{c}{\textbf{Telugu Perturbation Example (Character Removal, $p=50$)}} \\
\midrule

\textbf{Original:} \\
\foreignlanguage{telugu}{
సాహిత్య, విజ్ఞాన రంగాల్లో ముద్రణ ప్రక్రియకు కేంద్రస్థానం ఉంది. ఇది రచనలను విస్తరించడంలో అంశపూరిత పాత్ర పోషిస్తుంది. ప్రాచీన భారతదేశంలో ముఖ్యంగా హస్తప్రతులు ప్రాచుర్యంలో ఉండగా, యూరోపియన్ సాంప్రదాయాలలో ముద్రణ పద్దతులు వచ్చి విజ్ఞాన వ్యాప్తికి మౌలిక మద్దతునిచ్చాయి. 15వ శతాబ్దంలో గుటెన్బర్గ్ ప్రెస్ ఆవిష్కరణ తర్వాత పాఠ్య గ్రంథాలు ఎక్కువ సంఖ్యలో అందుబాటులోకి రాగా, అభివృద్ధి చెందిన దేశాల్లో విద్యారంగమే కాక సామాజిక సంస్కృతి కూడా ప్రగతిశీల మార్పులకు లోను అయింది. ఆధునిక కాలంలో పుస్తకాల ముద్రణ, అనువాద ప్రక్రియ అనుసంధానంతో మరింత విస్తృతమవుతోంది. వచనాలను ఒక భాషలోనుంచి మరో భాషలోకి మార్చడాన్ని సూచించే అనువాదం, జ్ఞానాభివృద్ధికి పెనుమార్గం చూపింది. సామాజిక, సాంస్కృతిక పరస్పర సంబంధాలను బలోపేతం చేయడంలో ఇది విశేష పాత్రను నిర్వర్తించింది. ఉన్నత విద్యాసంస్థల్లో, విశ్వవిద్యాలయ గ్రంథాలయం వంటి వనరుల ద్వారా ప్రయోజనాన్ని ఆనందించవచ్చు. విదేశీ రచనల జ్ఞానం అనువాద ప్రక్రియ కారణంగా సమాజానికి అందుతుంది. విద్యా విధానంలో మిషన్ ముఖ్యమైన ఆవశ్యకం. సాధికారత, విద్యార్థులకు నూతన ఆలోచనల్ని ఇవ్వడం వంటి లక్ష్యాలు విద్యా ప్రణాళికలో భాగంగా మారాయి. కొన్ని విద్యా సంస్థలు విశేష మిషన్ స్టేట్మెంట్లను రూపొందించి వాటిని ఆచరణలో పెట్టిన తీరు పరిశీలనార్హం. విద్యమార్గాన్ని విశాలంగా చేయడంలో, జ్ఞానం పంపిణీలో ఉపకరణాలుగా ముద్రిత గ్రంథాలు ఉపయోగపడతాయి. ఉపయోగకరమైన సమాచారాన్ని సమగ్రంగా అందించడంలో మిషన్ యొక్క ప్రాముఖ్యతను గుర్తించాలి. గ్రంథరచన ఒక సమగ్ర శాస్త్రీయ ప్రక్రియ. రచయితలు పరిశోధన, విమర్శన విశ్లేషణల ద్వారా సమాచారాన్ని సంకలనం చేస్తారు. విజ్ఞాన పరంగా, చిన్న పుస్తకం నుండి విశాల గ్రంధాలయ విలువైన వనరుల వరకు వివిధ స్థాయిలో రచనలు ప్రాచుర్యం పొందాయి. సాంప్రదాయము, ఆధునికత కలబోతతో రూపొందిన గ్రంథాలు సమాజానికి మార్గదర్శిగా నిలుస్తున్నాయి.
}

\vspace{0.4em}
\textbf{Perturbed:} \\
\bottomrule
\end{tabular}

\caption{Illustrative perturbation example created by random character deletion with probability $p=50$. Long-form samples are shown in a full-width layout to preserve readability for Indic scripts.}
\label{tab:perturbation_telugu}
\end{table*}

\begin{table*}[t]
\centering
\small
\setlength{\tabcolsep}{6pt}
\renewcommand{\arraystretch}{1.2}

\begin{tabular}{>{\RaggedRight\arraybackslash}p{0.97\textwidth}}
\toprule
\multicolumn{1}{c}{\textbf{Hindi Perturbation Example (Character Removal, $p=50$)}} \\
\midrule

\textbf{Original:} \\
\bottomrule
\end{tabular}

\caption{Illustrative perturbation example created by random character deletion with probability $p=50$. Long-form samples are shown in a full-width layout to preserve readability for Indic scripts.}
\label{tab:perturbation_hindi}
\end{table*}

\begin{table*}[t]
\centering
\small
\setlength{\tabcolsep}{6pt}
\renewcommand{\arraystretch}{1.2}

\begin{tabular}{>{\RaggedRight\arraybackslash}p{0.97\textwidth}}
\toprule
\multicolumn{1}{c}{\textbf{Tamil Perturbation Example (Character Removal, $p=50$)}} \\
\midrule

\textbf{Original:} \\
\TamilText{புவியியல் அறிவியல் துறையில் நிலத்தட்டுகளும், அவற்றைச் சூழவுள்ள இயக்கங்களும் மிக அதிக முக்கியத்துவம் பெறுகின்றன. நிலத்தட்டு என்பது பூமியின் மேற்பரப்பை மூடியிருக்கும் பெரும் பாறைத் தகடுகளாகும். இவை வெவ்வேறு பரப்பளவிலும், தடிமனுமாகவும் காணப்படுகின்றன. பூமியின் மேல்பகுதி மன்றாட் சூழல் போன்ற வெப்ப, அழுத்த வேறுபாடுகளினால், இவை சிறிது சிறிதாக நகர்கின்றன. இந்த நிலத்தட்டுகளுக்கிடையில் இருக்கும் எல்லைகள் வெடிப்புகள், இடர்பாடுகள், மோதல்கள் மற்றும் ஊடாடும் கருக்கைகளால் குறிப்பிட்ட இடங்களில் மிகுந்த அழுத்தம் ஏற்படுகிறது. சில நேரங்களில், நிலத்தட்டுகள் விரிவடையும் பகுதியில் அவை மாறுபாடு கொண்ட பதிப்புகளாக பிரிந்து போகும். இன்னும் சில நேரங்களில் தகடுகள் ஒன்றினொன்று மேல் செல்கின்றன அல்லது பாதியில் மோதுகின்றன. இந்த இயக்கங்கள் நிலைபெறவில்லை என்றால், திடீரென ஆற்றல் வெளிப்படுகிறது. இப்படியான ஆற்றல் வெளியீடு நிலநடுக்கத்தை உருவாக்குகிறது. நிலநடுக்கத்தின் போது பூமியின் மேற்பரப்பு பிசிறி, படிப்படியாக அதிர்ச்சி அலைகள் சுற்றியுள்ள பகுதிகளிலும் பரவுகின்றன. இது கட்டிடங்கள் மற்றும் பிற கட்டமைப்புகளில் சேதம் ஏற்படுத்த முடியும். டெக்டானிக் பிளேட் இயக்கங்கள் காரணமாக பெரும்பாலான நிலநடுக்கங்கள் ஏற்படுகின்றன. பூமியில் இருக்கும் இது போன்ற இயக்கங்கள் பெரும்பாலும் அகன்று ஆழத்தில் அல்லது புவியின் எல்லைகளில் நிகழ்கின்றன. மேலோட்டமான இயக்கங்கள் எழும்பும் பொழுது, சந்திக்கப்பட்ட இடத்தில் பெரும் அழுத்தத் தாக்கம் தெரிகிறது. சில சிக்கலான நிலைமைகளில், திடீரென ஆழமான பாகங்களில் நோடுகள் முறிவு பெறும்படி செய்கின்றன. இவ்வாறு நுழைந்தும் வெளியேறும் ஆற்றல் பூமியின் மேற்பரப்பை அதிர வைக்கும். சில நிலநடுக்கங்கள் கடலுக்கடியில் அல்லது குறுகிய ஆழத்தில் நிகழ்கின்றன. இவை கடல் நீரில் பெரும் அலைகள் அல்லது அதிர்ச்சியை உண்டாக்கும்.}

\vspace{0.4em}
\textbf{Perturbed:} \\
\bottomrule
\end{tabular}

\caption{Illustrative perturbation example created by random character deletion with probability $p=50$. Long-form samples are shown in a full-width layout to preserve readability for Indic scripts.}
\label{tab:perturbation_tamil}
\end{table*}

\begin{table*}[t]
\centering
\small
\setlength{\tabcolsep}{6pt}
\renewcommand{\arraystretch}{1.2}

\begin{tabular}{>{\RaggedRight\arraybackslash}p{0.97\textwidth}}
\toprule
\multicolumn{1}{c}{\textbf{Telugu Whitespace Attack Example ($\theta = 20.0\%$)}} \\
\midrule

\textbf{Original:} \\
\foreignlanguage{telugu}{
సాహిత్య, విజ్ఞాన రంగాల్లో ముద్రణ ప్రక్రియకు కేంద్రస్థానం ఉంది. ఇది రచనలను విస్తరించడంలో అంశపూరిత పాత్ర పోషిస్తుంది. ప్రాచీన భారతదేశంలో ముఖ్యంగా హస్తప్రతులు ప్రాచుర్యంలో ఉండగా, యూరోపియన్ సాంప్రదాయాలలో ముద్రణ పద్దతులు వచ్చి విజ్ఞాన వ్యాప్తికి మౌలిక మద్దతునిచ్చాయి. 15వ శతాబ్దంలో గుటెన్బర్గ్ ప్రెస్ ఆవిష్కరణ తర్వాత పాఠ్య గ్రంథాలు ఎక్కువ సంఖ్యలో అందుబాటులోకి రాగా, అభివృద్ధి చెందిన దేశాల్లో విద్యారంగమే కాక సామాజిక సంస్కృతి కూడా ప్రగతిశీల మార్పులకు లోను అయింది. ఆధునిక కాలంలో పుస్తకాల ముద్రణ, అనువాద ప్రక్రియ అనుసంధానంతో మరింత విస్తృతమవుతోంది. వచనాలను ఒక భాషలోనుంచి మరో భాషలోకి మార్చడాన్ని సూచించే అనువాదం, జ్ఞానాభివృద్ధికి పెనుమార్గం చూపింది. సామాజిక, సాంస్కృతిక పరస్పర సంబంధాలను బలోపేతం చేయడంలో ఇది విశేష పాత్రను నిర్వర్తించింది. ఉన్నత విద్యాసంస్థల్లో, విశ్వవిద్యాలయ గ్రంథాలయం వంటి వనరుల ద్వారా ప్రయోజనాన్ని ఆనందించవచ్చు. విదేశీ రచనల జ్ఞానం అనువాద ప్రక్రియ కారణంగా సమాజానికి అందుతుంది. విద్యా విధానంలో మిషన్ ముఖ్యమైన ఆవశ్యకం. సాధికారత, విద్యార్థులకు నూతన ఆలోచనల్ని ఇవ్వడం వంటి లక్ష్యాలు విద్యా ప్రణాళికలో భాగంగా మారాయి. కొన్ని విద్యా సంస్థలు విశేష మిషన్ స్టేట్మెంట్లను రూపొందించి వాటిని ఆచరణలో పెట్టిన తీరు పరిశీలనార్హం. విద్యమార్గాన్ని విశాలంగా చేయడంలో, జ్ఞానం పంపిణీలో ఉపకరణాలుగా ముద్రిత గ్రంథాలు ఉపయోగపడతాయి. ఉపయోగకరమైన సమాచారాన్ని సమగ్రంగా అందించడంలో మిషన్ యొక్క ప్రాముఖ్యతను గుర్తించాలి. గ్రంథరచన ఒక సమగ్ర శాస్త్రీయ ప్రక్రియ. రచయితలు పరిశోధన, విమర్శన విశ్లేషణల ద్వారా సమాచారాన్ని సంకలనం చేస్తారు. విజ్ఞాన పరంగా, చిన్న పుస్తకం నుండి విశాల గ్రంధాలయ విలువైన వనరుల వరకు వివిధ స్థాయిలో రచనలు ప్రాచుర్యం పొందాయి. సాంప్రదాయము, ఆధునికత కలబోతతో రూపొందిన గ్రంథాలు సమాజానికి మార్గదర్శిగా నిలుస్తున్నాయి.
}

\vspace{0.4em}
\textbf{Whitespace-Attacked:} \\
\bottomrule
\end{tabular}

\caption{Illustrative whitespace attack example for Telugu created by selectively increasing existing whitespace segments with modification rate $\theta = 20.0\%$. Long-form samples are shown in a full-width layout to preserve readability for Indic scripts.}
\label{tab:whitespace_telugu}
\end{table*}

\begin{table*}[t]
\centering
\small
\setlength{\tabcolsep}{6pt}
\renewcommand{\arraystretch}{1.2}

\begin{tabular}{>{\RaggedRight\arraybackslash}p{0.97\textwidth}}
\toprule
\multicolumn{1}{c}{\textbf{Hindi Whitespace Attack Example ($\theta = 20.0\%$)}} \\
\midrule

\textbf{Original:} \\
\bottomrule
\end{tabular}

\caption{Illustrative whitespace attack example for Hindi created by selectively increasing existing whitespace segments with modification rate $\theta = 20.0\%$. Long-form samples are shown in a full-width layout to preserve readability for Indic scripts.}
\label{tab:whitespace_hindi}
\end{table*}

\begin{table*}[t]
\centering
\small
\setlength{\tabcolsep}{6pt}
\renewcommand{\arraystretch}{1.2}

\begin{tabular}{>{\RaggedRight\arraybackslash}p{0.97\textwidth}}
\toprule
\multicolumn{1}{c}{\textbf{Tamil Whitespace Attack Example ($\theta = 20.0\%$)}} \\
\midrule

\textbf{Original:} \\
\TamilText{புவியியல் அறிவியல் துறையில் நிலத்தட்டுகளும், அவற்றைச் சூழவுள்ள இயக்கங்களும் மிக அதிக முக்கியத்துவம் பெறுகின்றன. நிலத்தட்டு என்பது பூமியின் மேற்பரப்பை மூடியிருக்கும் பெரும் பாறைத் தகடுகளாகும். இவை வெவ்வேறு பரப்பளவிலும், தடிமனுமாகவும் காணப்படுகின்றன. பூமியின் மேல்பகுதி மன்றாட் சூழல் போன்ற வெப்ப, அழுத்த வேறுபாடுகளினால், இவை சிறிது சிறிதாக நகர்கின்றன. இந்த நிலத்தட்டுகளுக்கிடையில் இருக்கும் எல்லைகள் வெடிப்புகள், இடர்பாடுகள், மோதல்கள் மற்றும் ஊடாடும் கருக்கைகளால் குறிப்பிட்ட இடங்களில் மிகுந்த அழுத்தம் ஏற்படுகிறது. சில நேரங்களில், நிலத்தட்டுகள் விரிவடையும் பகுதியில் அவை மாறுபாடு கொண்ட பதிப்புகளாக பிரிந்து போகும். இன்னும் சில நேரங்களில் தகடுகள் ஒன்றினொன்று மேல் செல்கின்றன அல்லது பாதியில் மோதுகின்றன. இந்த இயக்கங்கள் நிலைபெறவில்லை என்றால், திடீரென ஆற்றல் வெளிப்படுகிறது. இப்படியான ஆற்றல் வெளியீடு நிலநடுக்கத்தை உருவாக்குகிறது. நிலநடுக்கத்தின் போது பூமியின் மேற்பரப்பு பிசிறி, படிப்படியாக அதிர்ச்சி அலைகள் சுற்றியுள்ள பகுதிகளிலும் பரவுகின்றன. இது கட்டிடங்கள் மற்றும் பிற கட்டமைப்புகளில் சேதம் ஏற்படுத்த முடியும். டெக்டானிக் பிளேட் இயக்கங்கள் காரணமாக பெரும்பாலான நிலநடுக்கங்கள் ஏற்படுகின்றன. பூமியில் இருக்கும் இது போன்ற இயக்கங்கள் பெரும்பாலும் அகன்று ஆழத்தில் அல்லது புவியின் எல்லைகளில் நிகழ்கின்றன. மேலோட்டமான இயக்கங்கள் எழும்பும் பொழுது, சந்திக்கப்பட்ட இடத்தில் பெரும் அழுத்தத் தாக்கம் தெரிகிறது. சில சிக்கலான நிலைமைகளில், திடீரென ஆழமான பாகங்களில் நோடுகள் முறிவு பெறும்படி செய்கின்றன. இவ்வாறு நுழைந்தும் வெளியேறும் ஆற்றல் பூமியின் மேற்பரப்பை அதிர வைக்கும். சில நிலநடுக்கங்கள் கடலுக்கடியில் அல்லது குறுகிய ஆழத்தில் நிகழ்கின்றன. இவை கடல் நீரில் பெரும் அலைகள் அல்லது அதிர்ச்சியை உண்டாக்கும்.}

\vspace{0.4em}

\textbf{Whitespace-Attacked:} \\
\bottomrule
\end{tabular}

\caption{Illustrative whitespace attack example for Tamil created by selectively increasing existing whitespace segments with modification rate $\theta = 20.0\%$. Long-form samples are shown in a full-width layout to preserve readability for Indic scripts.}
\label{tab:whitespace_tamil}
\end{table*}

\begin{table*}[t]
\centering
\small
\setlength{\tabcolsep}{6pt}
\renewcommand{\arraystretch}{1.2}

\begin{tabular}{>{\RaggedRight\arraybackslash}p{0.97\textwidth}}
\toprule
\multicolumn{1}{c}{\textbf{Telugu Insert Paragraph Attack Example ($\theta = 50\%$)}} \\
\midrule

\textbf{Original:} \\
\foreignlanguage{telugu}{
సాహిత్య, విజ్ఞాన రంగాల్లో ముద్రణ ప్రక్రియకు కేంద్రస్థానం ఉంది. ఇది రచనలను విస్తరించడంలో అంశపూరిత పాత్ర పోషిస్తుంది. ప్రాచీన భారతదేశంలో ముఖ్యంగా హస్తప్రతులు ప్రాచుర్యంలో ఉండగా, యూరోపియన్ సాంప్రదాయాలలో ముద్రణ పద్దతులు వచ్చి విజ్ఞాన వ్యాప్తికి మౌలిక మద్దతునిచ్చాయి. 15వ శతాబ్దంలో గుటెన్బర్గ్ ప్రెస్ ఆవిష్కరణ తర్వాత పాఠ్య గ్రంథాలు ఎక్కువ సంఖ్యలో అందుబాటులోకి రాగా, అభివృద్ధి చెందిన దేశాల్లో విద్యారంగమే కాక సామాజిక సంస్కృతి కూడా ప్రగతిశీల మార్పులకు లోను అయింది. ఆధునిక కాలంలో పుస్తకాల ముద్రణ, అనువాద ప్రక్రియ అనుసంధానంతో మరింత విస్తృతమవుతోంది. వచనాలను ఒక భాషలోనుంచి మరో భాషలోకి మార్చడాన్ని సూచించే అనువాదం, జ్ఞానాభివృద్ధికి పెనుమార్గం చూపింది. సామాజిక, సాంస్కృతిక పరస్పర సంబంధాలను బలోపేతం చేయడంలో ఇది విశేష పాత్రను నిర్వర్తించింది. ఉన్నత విద్యాసంస్థల్లో, విశ్వవిద్యాలయ గ్రంథాలయం వంటి వనరుల ద్వారా ప్రయోజనాన్ని ఆనందించవచ్చు. విదేశీ రచనల జ్ఞానం అనువాద ప్రక్రియ కారణంగా సమాజానికి అందుతుంది. విద్యా విధానంలో మిషన్ ముఖ్యమైన ఆవశ్యకం. సాధికారత, విద్యార్థులకు నూతన ఆలోచనల్ని ఇవ్వడం వంటి లక్ష్యాలు విద్యా ప్రణాళికలో భాగంగా మారాయి. కొన్ని విద్యా సంస్థలు విశేష మిషన్ స్టేట్మెంట్లను రూపొందించి వాటిని ఆచరణలో పెట్టిన తీరు పరిశీలనార్హం. విద్యమార్గాన్ని విశాలంగా చేయడంలో, జ్ఞానం పంపిణీలో ఉపకరణాలుగా ముద్రిత గ్రంథాలు ఉపయోగపడతాయి. ఉపయోగకరమైన సమాచారాన్ని సమగ్రంగా అందించడంలో మిషన్ యొక్క ప్రాముఖ్యతను గుర్తించాలి. గ్రంథరచన ఒక సమగ్ర శాస్త్రీయ ప్రక్రియ. రచయితలు పరిశోధన, విమర్శన విశ్లేషణల ద్వారా సమాచారాన్ని సంకలనం చేస్తారు. విజ్ఞాన పరంగా, చిన్న పుస్తకం నుండి విశాల గ్రంధాలయ విలువైన వనరుల వరకు వివిధ స్థాయిలో రచనలు ప్రాచుర్యం పొందాయి. సాంప్రదాయము, ఆధునికత కలబోతతో రూపొందిన గ్రంథాలు సమాజానికి మార్గదర్శిగా నిలుస్తున్నాయి.
}

\vspace{0.4em}
\textbf{Insert-Paragraph-Attacked:} \\
\bottomrule
\end{tabular}

\caption{Illustrative insert paragraph attack example for Telugu created by inserting paragraph breaks at selected sentence boundaries with insertion rate $\theta = 50\%$. Long-form samples are shown in a full-width layout to preserve readability for Indic scripts.}
\label{tab:insert_paragraph_telugu}
\end{table*}

\begin{table*}[t]
\centering
\small
\setlength{\tabcolsep}{6pt}
\renewcommand{\arraystretch}{1.2}

\begin{tabular}{>{\RaggedRight\arraybackslash}p{0.97\textwidth}}
\toprule
\multicolumn{1}{c}{\textbf{Hindi Insert Paragraph Attack Example ($\theta = 50\%$)}} \\
\midrule

\textbf{Original:} \\
\bottomrule
\end{tabular}

\caption{Illustrative insert paragraph attack example for Hindi created by inserting paragraph breaks at selected sentence boundaries with insertion rate $\theta = 50\%$. Long-form samples are shown in a full-width layout to preserve readability for Indic scripts.}
\label{tab:insert_paragraph_hindi}
\end{table*}
\begin{table*}[t]
\centering
\small
\setlength{\tabcolsep}{6pt}
\renewcommand{\arraystretch}{1.2}

\begin{tabular}{>{\RaggedRight\arraybackslash}p{0.97\textwidth}}
\toprule
\multicolumn{1}{c}{\textbf{Tamil Insert Paragraph Attack Example ($\theta = 50\%$)}} \\
\midrule

\textbf{Original:} \\
\TamilText{புவியியல் அறிவியல் துறையில் நிலத்தட்டுகளும், அவற்றைச் சூழவுள்ள இயக்கங்களும் மிக அதிக முக்கியத்துவம் பெறுகின்றன. நிலத்தட்டு என்பது பூமியின் மேற்பரப்பை மூடியிருக்கும் பெரும் பாறைத் தகடுகளாகும். இவை வெவ்வேறு பரப்பளவிலும், தடிமனுமாகவும் காணப்படுகின்றன. பூமியின் மேல்பகுதி மன்றாட் சூழல் போன்ற வெப்ப, அழுத்த வேறுபாடுகளினால், இவை சிறிது சிறிதாக நகர்கின்றன. இந்த நிலத்தட்டுகளுக்கிடையில் இருக்கும் எல்லைகள் வெடிப்புகள், இடர்பாடுகள், மோதல்கள் மற்றும் ஊடாடும் கருக்கைகளால் குறிப்பிட்ட இடங்களில் மிகுந்த அழுத்தம் ஏற்படுகிறது. சில நேரங்களில், நிலத்தட்டுகள் விரிவடையும் பகுதியில் அவை மாறுபாடு கொண்ட பதிப்புகளாக பிரிந்து போகும். இன்னும் சில நேரங்களில் தகடுகள் ஒன்றினொன்று மேல் செல்கின்றன அல்லது பாதியில் மோதுகின்றன. இந்த இயக்கங்கள் நிலைபெறவில்லை என்றால், திடீரென ஆற்றல் வெளிப்படுகிறது. இப்படியான ஆற்றல் வெளியீடு நிலநடுக்கத்தை உருவாக்குகிறது. நிலநடுக்கத்தின் போது பூமியின் மேற்பரப்பு பிசிறி, படிப்படியாக அதிர்ச்சி அலைகள் சுற்றியுள்ள பகுதிகளிலும் பரவுகின்றன. இது கட்டிடங்கள் மற்றும் பிற கட்டமைப்புகளில் சேதம் ஏற்படுத்த முடியும். டெக்டானிக் பிளேட் இயக்கங்கள் காரணமாக பெரும்பாலான நிலநடுக்கங்கள் ஏற்படுகின்றன. பூமியில் இருக்கும் இது போன்ற இயக்கங்கள் பெரும்பாலும் அகன்று ஆழத்தில் அல்லது புவியின் எல்லைகளில் நிகழ்கின்றன. மேலோட்டமான இயக்கங்கள் எழும்பும் பொழுது, சந்திக்கப்பட்ட இடத்தில் பெரும் அழுத்தத் தாக்கம் தெரிகிறது. சில சிக்கலான நிலைமைகளில், திடீரென ஆழமான பாகங்களில் நோடுகள் முறிவு பெறும்படி செய்கின்றன. இவ்வாறு நுழைந்தும் வெளியேறும் ஆற்றல் பூமியின் மேற்பரப்பை அதிர வைக்கும். சில நிலநடுக்கங்கள் கடலுக்கடியில் அல்லது குறுகிய ஆழத்தில் நிகழ்கின்றன. இவை கடல் நீரில் பெரும் அலைகள் அல்லது அதிர்ச்சியை உண்டாக்கும்.}

\vspace{0.4em}
\textbf{Insert-Paragraph-Attacked:} \\
\bottomrule
\end{tabular}

\caption{Illustrative insert paragraph attack example for Tamil created by inserting paragraph breaks at selected sentence boundaries with insertion rate $\theta = 50\%$. Long-form samples are shown in a full-width layout to preserve readability for Indic scripts.}
\label{tab:insert_paragraph_tamil}
\end{table*}
\begin{table*}[t]
\centering
\small
\setlength{\tabcolsep}{6pt}
\renewcommand{\arraystretch}{1.2}

\begin{tabular}{>{\RaggedRight\arraybackslash}p{0.97\textwidth}}
\toprule
\multicolumn{1}{c}{\textbf{Telugu Alternative Spelling Attack Example ($p = 1.0$)}} \\
\midrule

\textbf{Original:} \\
\foreignlanguage{telugu}{
సాహిత్య, విజ్ఞాన రంగాల్లో ముద్రణ ప్రక్రియకు కేంద్రస్థానం ఉంది. ఇది రచనలను విస్తరించడంలో అంశపూరిత పాత్ర పోషిస్తుంది. ప్రాచీన భారతదేశంలో ముఖ్యంగా హస్తప్రతులు ప్రాచుర్యంలో ఉండగా, యూరోపియన్ సాంప్రదాయాలలో ముద్రణ పద్దతులు వచ్చి విజ్ఞాన వ్యాప్తికి మౌలిక మద్దతునిచ్చాయి. 15వ శతాబ్దంలో గుటెన్బర్గ్ ప్రెస్ ఆవిష్కరణ తర్వాత పాఠ్య గ్రంథాలు ఎక్కువ సంఖ్యలో అందుబాటులోకి రాగా, అభివృద్ధి చెందిన దేశాల్లో విద్యారంగమే కాక సామాజిక సంస్కృతి కూడా ప్రగతిశీల మార్పులకు లోను అయింది. ఆధునిక కాలంలో పుస్తకాల ముద్రణ, అనువాద ప్రక్రియ అనుసంధానంతో మరింత విస్తృతమవుతోంది. వచనాలను ఒక భాషలోనుంచి మరో భాషలోకి మార్చడాన్ని సూచించే అనువాదం, జ్ఞానాభివృద్ధికి పెనుమార్గం చూపింది. సామాజిక, సాంస్కృతిక పరస్పర సంబంధాలను బలోపేతం చేయడంలో ఇది విశేష పాత్రను నిర్వర్తించింది. ఉన్నత విద్యాసంస్థల్లో, విశ్వవిద్యాలయ గ్రంథాలయం వంటి వనరుల ద్వారా ప్రయోజనాన్ని ఆనందించవచ్చు. విదేశీ రచనల జ్ఞానం అనువాద ప్రక్రియ కారణంగా సమాజానికి అందుతుంది. విద్యా విధానంలో మిషన్ ముఖ్యమైన ఆవశ్యకం. సాధికారత, విద్యార్థులకు నూతన ఆలోచనల్ని ఇవ్వడం వంటి లక్ష్యాలు విద్యా ప్రణాళికలో భాగంగా మారాయి. కొన్ని విద్యా సంస్థలు విశేష మిషన్ స్టేట్మెంట్లను రూపొందించి వాటిని ఆచరణలో పెట్టిన తీరు పరిశీలనార్హం. విద్యమార్గాన్ని విశాలంగా చేయడంలో, జ్ఞానం పంపిణీలో ఉపకరణాలుగా ముద్రిత గ్రంథాలు ఉపయోగపడతాయి. ఉపయోగకరమైన సమాచారాన్ని సమగ్రంగా అందించడంలో మిషన్ యొక్క ప్రాముఖ్యతను గుర్తించాలి. గ్రంథరచన ఒక సమగ్ర శాస్త్రీయ ప్రక్రియ. రచయితలు పరిశోధన, విమర్శన విశ్లేషణల ద్వారా సమాచారాన్ని సంకలనం చేస్తారు. విజ్ఞాన పరంగా, చిన్న పుస్తకం నుండి విశాల గ్రంధాలయ విలువైన వనరుల వరకు వివిధ స్థాయిలో రచనలు ప్రాచుర్యం పొందాయి. సాంప్రదాయము, ఆధునికత కలబోతతో రూపొందిన గ్రంథాలు సమాజానికి మార్గదర్శిగా నిలుస్తున్నాయి.
}

\vspace{0.4em}
\textbf{Alternative-Spelling-Attacked:} \\
\bottomrule
\end{tabular}

\caption{Illustrative alternative spelling attack example for Telugu created by replacing dictionary-covered words with valid orthographic variants using replacement probability $p = 1.0$.Alternative spelling attack segments are shown in \textbf{bold} for clarity.}
\label{tab:alt_spelling_telugu}
\end{table*}

\begin{table*}[t]
\centering
\small
\setlength{\tabcolsep}{6pt}
\renewcommand{\arraystretch}{1.2}

\begin{tabular}{>{\RaggedRight\arraybackslash}p{0.97\textwidth}}
\toprule
\multicolumn{1}{c}{\textbf{Hindi Alternative Spelling Attack Example ($p = 1.0$)}} \\
\midrule

\textbf{Original:} \\
\bottomrule
\end{tabular}

\caption{Illustrative alternative spelling attack example for Hindi created by replacing dictionary-covered words with valid orthographic variants using replacement probability $p = 1.0$.Alternative spelling attack segments are shown in \textbf{bold} for clarity.}
\label{tab:alt_spelling_hindi}
\end{table*}
\begin{table*}[t]
\centering
\small
\setlength{\tabcolsep}{6pt}
\renewcommand{\arraystretch}{1.2}

\begin{tabular}{>{\RaggedRight\arraybackslash}p{0.97\textwidth}}
\toprule
\multicolumn{1}{c}{\textbf{Tamil Alternative Spelling Attack Example ($p = 1.0$)}} \\
\midrule

\textbf{Original:} \\
\TamilText{
புவியியல் அறிவியல் துறையில் நிலத்தட்டுகளும், அவற்றைச் சூழவுள்ள இயக்கங்களும் மிக அதிக முக்கியத்துவம் பெறுகின்றன. நிலத்தட்டு என்பது பூமியின் மேற்பரப்பை மூடியிருக்கும் பெரும் பாறைத் தகடுகளாகும். இவை வெவ்வேறு பரப்பளவிலும், தடிமனுமாகவும் காணப்படுகின்றன. பூமியின் மேல்பகுதி மன்றாட் சூழல் போன்ற வெப்ப, அழுத்த வேறுபாடுகளினால், இவை சிறிது சிறிதாக நகர்கின்றன. இந்த நிலத்தட்டுகளுக்கிடையில் இருக்கும் எல்லைகள் வெடிப்புகள், இடர்பாடுகள், மோதல்கள் மற்றும் ஊடாடும் கருக்கைகளால் குறிப்பிட்ட இடங்களில் மிகுந்த அழுத்தம் ஏற்படுகிறது. சில நேரங்களில், நிலத்தட்டுகள் விரிவடையும் பகுதியில் அவை மாறுபாடு கொண்ட பதிப்புகளாக பிரிந்து போகும். இன்னும் சில நேரங்களில் தகடுகள் ஒன்றினொன்று மேல் செல்கின்றன அல்லது பாதியில் மோதுகின்றன. இந்த இயக்கங்கள் நிலைபெறவில்லை என்றால், திடீரென ஆற்றல் வெளிப்படுகிறது. இப்படியான ஆற்றல் வெளியீடு நிலநடுக்கத்தை உருவாக்குகிறது. நிலநடுக்கத்தின் போது பூமியின் மேற்பரப்பு பிசிறி, படிப்படியாக அதிர்ச்சி அலைகள் சுற்றியுள்ள பகுதிகளிலும் பரவுகின்றன. இது கட்டிடங்கள் மற்றும் பிற கட்டமைப்புகளில் சேதம் ஏற்படுத்த முடியும். டெக்டானிக் பிளேட் இயக்கங்கள் காரணமாக பெரும்பாலான நிலநடுக்கங்கள் ஏற்படுகின்றன. பூமியில் இருக்கும் இது போன்ற இயக்கங்கள் பெரும்பாலும் அகன்று ஆழத்தில் அல்லது புவியின் எல்லைகளில் நிகழ்கின்றன. மேலோட்டமான இயக்கங்கள் எழும்பும் பொழுது, சந்திக்கப்பட்ட இடத்தில் பெரும் அழுத்தத் தாக்கம் தெரிகிறது. சில சிக்கலான நிலைமைகளில், திடீரென ஆழமான பாகங்களில் நோடுகள் முறிவு பெறும்படி செய்கின்றன. இவ்வாறு நுழைந்தும் வெளியேறும் ஆற்றல் பூமியின் மேற்பரப்பை அதிர வைக்கும். சில நிலநடுக்கங்கள் கடலுக்கடியில் அல்லது குறுகிய ஆழத்தில் நிகழ்கின்றன. இவை கடல் நீரில் பெரும் அலைகள் அல்லது அதிர்ச்சியை உண்டாக்கும்.
}

\vspace{0.4em}
\textbf{Alternative-Spelling-Attacked:} \\
\bottomrule
\end{tabular}

\caption{Illustrative alternative spelling attack example for Tamil created by replacing dictionary-covered words with valid orthographic variants using replacement probability $p = 1.0$.Alternative spelling attack segments are shown in \textbf{bold} for clarity.}
\label{tab:alt_spelling_tamil}
\end{table*}
\begin{table*}[t]
\centering
\small
\setlength{\tabcolsep}{6pt}
\renewcommand{\arraystretch}{1.2}

\begin{tabular}{>{\RaggedRight\arraybackslash}p{0.97\textwidth}}
\toprule
\multicolumn{1}{c}{\textbf{Telugu Misspelling Attack Example ($p = 1.0$)}} \\
\midrule

\textbf{Original:} \\
\foreignlanguage{telugu}{
సాహిత్య, విజ్ఞాన రంగాల్లో ముద్రణ ప్రక్రియకు కేంద్రస్థానం ఉంది. ఇది రచనలను విస్తరించడంలో అంశపూరిత పాత్ర పోషిస్తుంది. ప్రాచీన భారతదేశంలో ముఖ్యంగా హస్తప్రతులు ప్రాచుర్యంలో ఉండగా, యూరోపియన్ సాంప్రదాయాలలో ముద్రణ పద్దతులు వచ్చి విజ్ఞాన వ్యాప్తికి మౌలిక మద్దతునిచ్చాయి. 15వ శతాబ్దంలో గుటెన్బర్గ్ ప్రెస్ ఆవిష్కరణ తర్వాత పాఠ్య గ్రంథాలు ఎక్కువ సంఖ్యలో అందుబాటులోకి రాగా, అభివృద్ధి చెందిన దేశాల్లో విద్యారంగమే కాక సామాజిక సంస్కృతి కూడా ప్రగతిశీల మార్పులకు లోను అయింది. ఆధునిక కాలంలో పుస్తకాల ముద్రణ, అనువాద ప్రక్రియ అనుసంధానంతో మరింత విస్తృతమవుతోంది. వచనాలను ఒక భాషలోనుంచి మరో భాషలోకి మార్చడాన్ని సూచించే అనువాదం, జ్ఞానాభివృద్ధికి పెనుమార్గం చూపింది. సామాజిక, సాంస్కృతిక పరస్పర సంబంధాలను బలోపేతం చేయడంలో ఇది విశేష పాత్రను నిర్వర్తించింది. ఉన్నత విద్యాసంస్థల్లో, విశ్వవిద్యాలయ గ్రంథాలయం వంటి వనరుల ద్వారా ప్రయోజనాన్ని ఆనందించవచ్చు. విదేశీ రచనల జ్ఞానం అనువాద ప్రక్రియ కారణంగా సమాజానికి అందుతుంది. విద్యా విధానంలో మిషన్ ముఖ్యమైన ఆవశ్యకం. సాధికారత, విద్యార్థులకు నూతన ఆలోచనల్ని ఇవ్వడం వంటి లక్ష్యాలు విద్యా ప్రణాళికలో భాగంగా మారాయి. కొన్ని విద్యా సంస్థలు విశేష మిషన్ స్టేట్మెంట్లను రూపొందించి వాటిని ఆచరణలో పెట్టిన తీరు పరిశీలనార్హం. విద్యమార్గాన్ని విశాలంగా చేయడంలో, జ్ఞానం పంపిణీలో ఉపకరణాలుగా ముద్రిత గ్రంథాలు ఉపయోగపడతాయి. ఉపయోగకరమైన సమాచారాన్ని సమగ్రంగా అందించడంలో మిషన్ యొక్క ప్రాముఖ్యతను గుర్తించాలి. గ్రంథరచన ఒక సమగ్ర శాస్త్రీయ ప్రక్రియ. రచయితలు పరిశోధన, విమర్శన విశ్లేషణల ద్వారా సమాచారాన్ని సంకలనం చేస్తారు. విజ్ఞాన పరంగా, చిన్న పుస్తకం నుండి విశాల గ్రంధాలయ విలువైన వనరుల వరకు వివిధ స్థాయిలో రచనలు ప్రాచుర్యం పొందాయి. సాంప్రదాయము, ఆధునికత కలబోతతో రూపొందిన గ్రంథాలు సమాజానికి మార్గదర్శిగా నిలుస్తున్నాయి.
}

\vspace{0.4em}
\textbf{Misspelling-Attacked:} \\
\bottomrule
\end{tabular}

\caption{Illustrative misspelling attack example for Telugu created by replacing dictionary-covered words with common typographical errors using replacement probability $p = 1.0$.Segments containing misspelling perturbations are highlighted in \textbf{bold} for ease of identification.}
\label{tab:misspelling_telugu}
\end{table*}

\begin{table*}[t]
\centering
\small
\setlength{\tabcolsep}{6pt}
\renewcommand{\arraystretch}{1.2}

\begin{tabular}{>{\RaggedRight\arraybackslash}p{0.97\textwidth}}
\toprule
\multicolumn{1}{c}{\textbf{Hindi Misspelling Attack Example ($p = 1.0$)}} \\
\midrule

\textbf{Original:} \\
\bottomrule
\end{tabular}

\caption{Illustrative misspelling attack example for Hindi created by replacing dictionary-covered words with common typographical errors using replacement probability $p = 1.0$.Segments containing misspelling perturbations are highlighted in \textbf{bold} for ease of identification.}
\label{tab:misspelling_hindi}
\end{table*}

\begin{table*}[t]
\centering
\small
\setlength{\tabcolsep}{6pt}
\renewcommand{\arraystretch}{1.2}

\begin{tabular}{>{\RaggedRight\arraybackslash}p{0.97\textwidth}}
\toprule
\multicolumn{1}{c}{\textbf{Tamil Misspelling Attack Example ($p = 1.0$)}} \\
\midrule

\textbf{Original:} \\
\TamilText{
புவியியல் அறிவியல் துறையில் நிலத்தட்டுகளும், அவற்றைச் சூழவுள்ள இயக்கங்களும் மிக அதிக முக்கியத்துவம் பெறுகின்றன. நிலத்தட்டு என்பது பூமியின் மேற்பரப்பை மூடியிருக்கும் பெரும் பாறைத் தகடுகளாகும். இவை வெவ்வேறு பரப்பளவிலும், தடிமனுமாகவும் காணப்படுகின்றன. பூமியின் மேல்பகுதி மன்றாட் சூழல் போன்ற வெப்ப, அழுத்த வேறுபாடுகளினால், இவை சிறிது சிறிதாக நகர்கின்றன. இந்த நிலத்தட்டுகளுக்கிடையில் இருக்கும் எல்லைகள் வெடிப்புகள், இடர்பாடுகள், மோதல்கள் மற்றும் ஊடாடும் கருக்கைகளால் குறிப்பிட்ட இடங்களில் மிகுந்த அழுத்தம் ஏற்படுகிறது. சில நேரங்களில், நிலத்தட்டுகள் விரிவடையும் பகுதியில் அவை மாறுபாடு கொண்ட பதிப்புகளாக பிரிந்து போகும். இன்னும் சில நேரங்களில் தகடுகள் ஒன்றினொன்று மேல் செல்கின்றன அல்லது பாதியில் மோதுகின்றன. இந்த இயக்கங்கள் நிலைபெறவில்லை என்றால், திடீரென ஆற்றல் வெளிப்படுகிறது. இப்படியான ஆற்றல் வெளியீடு நிலநடுக்கத்தை உருவாக்குகிறது. நிலநடுக்கத்தின் போது பூமியின் மேற்பரப்பு பிசிறி, படிப்படியாக அதிர்ச்சி அலைகள் சுற்றியுள்ள பகுதிகளிலும் பரவுகின்றன. இது கட்டிடங்கள் மற்றும் பிற கட்டமைப்புகளில் சேதம் ஏற்படுத்த முடியும். டெக்டானிக் பிளேட் இயக்கங்கள் காரணமாக பெரும்பாலான நிலநடுக்கங்கள் ஏற்படுகின்றன. பூமியில் இருக்கும் இது போன்ற இயக்கங்கள் பெரும்பாலும் அகன்று ஆழத்தில் அல்லது புவியின் எல்லைகளில் நிகழ்கின்றன. மேலோட்டமான இயக்கங்கள் எழும்பும் பொழுது, சந்திக்கப்பட்ட இடத்தில் பெரும் அழுத்தத் தாக்கம் தெரிகிறது. சில சிக்கலான நிலைமைகளில், திடீரென ஆழமான பாகங்களில் நோடுகள் முறிவு பெறும்படி செய்கின்றன. இவ்வாறு நுழைந்தும் வெளியேறும் ஆற்றல் பூமியின் மேற்பரப்பை அதிர வைக்கும். சில நிலநடுக்கங்கள் கடலுக்கடியில் அல்லது குறுகிய ஆழத்தில் நிகழ்கின்றன. இவை கடல் நீரில் பெரும் அலைகள் அல்லது அதிர்ச்சியை உண்டாக்கும்.
}

\vspace{0.4em}
\textbf{Misspelling-Attacked:} \\
\bottomrule
\end{tabular}

\caption{Illustrative misspelling attack example for Tamil created by replacing dictionary-covered words with common typographical errors using replacement probability $p = 1.0$.Segments containing misspelling perturbations are highlighted in \textbf{bold} for ease of identification.}
\label{tab:misspelling_tamil}
\end{table*}
\begin{table*}[t]
\centering
\small
\setlength{\tabcolsep}{6pt}
\renewcommand{\arraystretch}{1.2}

\begin{tabular}{>{\RaggedRight\arraybackslash}p{0.97\textwidth}}
\toprule
\multicolumn{1}{c}{\textbf{Telugu Synonym Swap Attack Example ($p = 1.0$)}} \\
\midrule

\textbf{Original:} \\
\foreignlanguage{telugu}{
సాహిత్య, విజ్ఞాన రంగాల్లో ముద్రణ ప్రక్రియకు కేంద్రస్థానం ఉంది. ఇది రచనలను విస్తరించడంలో అంశపూరిత పాత్ర పోషిస్తుంది. ప్రాచీన భారతదేశంలో ముఖ్యంగా హస్తప్రతులు ప్రాచుర్యంలో ఉండగా, యూరోపియన్ సాంప్రదాయాలలో ముద్రణ పద్దతులు వచ్చి విజ్ఞాన వ్యాప్తికి మౌలిక మద్దతునిచ్చాయి. 15వ శతాబ్దంలో గుటెన్బర్గ్ ప్రెస్ ఆవిష్కరణ తర్వాత పాఠ్య గ్రంథాలు ఎక్కువ సంఖ్యలో అందుబాటులోకి రాగా, అభివృద్ధి చెందిన దేశాల్లో విద్యారంగమే కాక సామాజిక సంస్కృతి కూడా ప్రగతిశీల మార్పులకు లోను అయింది. ఆధునిక కాలంలో పుస్తకాల ముద్రణ, అనువాద ప్రక్రియ అనుసంధానంతో మరింత విస్తృతమవుతోంది. వచనాలను ఒక భాషలోనుంచి మరో భాషలోకి మార్చడాన్ని సూచించే అనువాదం, జ్ఞానాభివృద్ధికి పెనుమార్గం చూపింది. సామాజిక, సాంస్కృతిక పరస్పర సంబంధాలను బలోపేతం చేయడంలో ఇది విశేష పాత్రను నిర్వర్తించింది. ఉన్నత విద్యాసంస్థల్లో, విశ్వవిద్యాలయ గ్రంథాలయం వంటి వనరుల ద్వారా ప్రయోజనాన్ని ఆనందించవచ్చు. విదేశీ రచనల జ్ఞానం అనువాద ప్రక్రియ కారణంగా సమాజానికి అందుతుంది. విద్యా విధానంలో మిషన్ ముఖ్యమైన ఆవశ్యకం. సాధికారత, విద్యార్థులకు నూతన ఆలోచనల్ని ఇవ్వడం వంటి లక్ష్యాలు విద్యా ప్రణాళికలో భాగంగా మారాయి. కొన్ని విద్యా సంస్థలు విశేష మిషన్ స్టేట్మెంట్లను రూపొందించి వాటిని ఆచరణలో పెట్టిన తీరు పరిశీలనార్హం. విద్యమార్గాన్ని విశాలంగా చేయడంలో, జ్ఞానం పంపిణీలో ఉపకరణాలుగా ముద్రిత గ్రంథాలు ఉపయోగపడతాయి. ఉపయోగకరమైన సమాచారాన్ని సమగ్రంగా అందించడంలో మిషన్ యొక్క ప్రాముఖ్యతను గుర్తించాలి. గ్రంథరచన ఒక సమగ్ర శాస్త్రీయ ప్రక్రియ. రచయితలు పరిశోధన, విమర్శన విశ్లేషణల ద్వారా సమాచారాన్ని సంకలనం చేస్తారు. విజ్ఞాన పరంగా, చిన్న పుస్తకం నుండి విశాల గ్రంధాలయ విలువైన వనరుల వరకు వివిధ స్థాయిలో రచనలు ప్రాచుర్యం పొందాయి. సాంప్రదాయము, ఆధునికత కలబోతతో రూపొందిన గ్రంథాలు సమాజానికి మార్గదర్శిగా నిలుస్తున్నాయి.
}

\vspace{0.4em}
\textbf{Synonym-Swapped:} \\
\bottomrule
\end{tabular}

\caption{Illustrative synonym swap attack example for Telugu created by replacing dictionary-covered words with meaning-preserving synonyms using replacement probability $p = 1.0$.Segments containing synonym swap perturbations are highlighted in \textbf{bold} for ease of identification.}
\label{tab:synonym_swap_telugu}
\end{table*}

\begin{table*}[t]
\centering
\small
\setlength{\tabcolsep}{6pt}
\renewcommand{\arraystretch}{1.2}

\begin{tabular}{>{\RaggedRight\arraybackslash}p{0.97\textwidth}}
\toprule
\multicolumn{1}{c}{\textbf{Hindi Synonym Swap Attack Example ($p = 1.0$)}} \\
\midrule

\textbf{Original:} \\
\bottomrule
\end{tabular}

\caption{Illustrative synonym swap attack example for Hindi created by replacing dictionary-covered words with meaning-preserving synonyms using replacement probability $p = 1.0$.Segments containing synonym swap perturbations are highlighted in \textbf{bold} for ease of identification.}
\label{tab:synonym_swap_hindi}
\end{table*}

\begin{table*}[t]
\centering
\small
\setlength{\tabcolsep}{6pt}
\renewcommand{\arraystretch}{1.2}

\begin{tabular}{>{\RaggedRight\arraybackslash}p{0.97\textwidth}}
\toprule
\multicolumn{1}{c}{\textbf{Tamil Synonym Swap Attack Example ($p = 1.0$)}} \\
\midrule

\textbf{Original:} \\
\TamilText{
புவியியல் அறிவியல் துறையில் நிலத்தட்டுகளும், அவற்றைச் சூழவுள்ள இயக்கங்களும் மிக அதிக முக்கியத்துவம் பெறுகின்றன. நிலத்தட்டு என்பது பூமியின் மேற்பரப்பை மூடியிருக்கும் பெரும் பாறைத் தகடுகளாகும். இவை வெவ்வேறு பரப்பளவிலும், தடிமனுமாகவும் காணப்படுகின்றன. பூமியின் மேல்பகுதி மன்றாட் சூழல் போன்ற வெப்ப, அழுத்த வேறுபாடுகளினால், இவை சிறிது சிறிதாக நகர்கின்றன. இந்த நிலத்தட்டுகளுக்கிடையில் இருக்கும் எல்லைகள் வெடிப்புகள், இடர்பாடுகள், மோதல்கள் மற்றும் ஊடாடும் கருக்கைகளால் குறிப்பிட்ட இடங்களில் மிகுந்த அழுத்தம் ஏற்படுகிறது. சில நேரங்களில், நிலத்தட்டுகள் விரிவடையும் பகுதியில் அவை மாறுபாடு கொண்ட பதிப்புகளாக பிரிந்து போகும். இன்னும் சில நேரங்களில் தகடுகள் ஒன்றினொன்று மேல் செல்கின்றன அல்லது பாதியில் மோதுகின்றன. இந்த இயக்கங்கள் நிலைபெறவில்லை என்றால், திடீரென ஆற்றல் வெளிப்படுகிறது. இப்படியான ஆற்றல் வெளியீடு நிலநடுக்கத்தை உருவாக்குகிறது. நிலநடுக்கத்தின் போது பூமியின் மேற்பரப்பு பிசிறி, படிப்படியாக அதிர்ச்சி அலைகள் சுற்றியுள்ள பகுதிகளிலும் பரவுகின்றன. இது கட்டிடங்கள் மற்றும் பிற கட்டமைப்புகளில் சேதம் ஏற்படுத்த முடியும். டெக்டானிக் பிளேட் இயக்கங்கள் காரணமாக பெரும்பாலான நிலநடுக்கங்கள் ஏற்படுகின்றன. பூமியில் இருக்கும் இது போன்ற இயக்கங்கள் பெரும்பாலும் அகன்று ஆழத்தில் அல்லது புவியின் எல்லைகளில் நிகழ்கின்றன. மேலோட்டமான இயக்கங்கள் எழும்பும் பொழுது, சந்திக்கப்பட்ட இடத்தில் பெரும் அழுத்தத் தாக்கம் தெரிகிறது. சில சிக்கலான நிலைமைகளில், திடீரென ஆழமான பாகங்களில் நோடுகள் முறிவு பெறும்படி செய்கின்றன. இவ்வாறு நுழைந்தும் வெளியேறும் ஆற்றல் பூமியின் மேற்பரப்பை அதிர வைக்கும். சில நிலநடுக்கங்கள் கடலுக்கடியில் அல்லது குறுகிய ஆழத்தில் நிகழ்கின்றன. இவை கடல் நீரில் பெரும் அலைகள் அல்லது அதிர்ச்சியை உண்டாக்கும்.
}

\vspace{0.4em}
\textbf{Synonym-Swapped:} \\
\bottomrule
\end{tabular}

\caption{Illustrative synonym swap attack example for Tamil created by replacing dictionary-covered words with meaning-preserving synonyms using replacement probability $p = 1.0$.Segments containing synonym swap perturbations are highlighted in \textbf{bold} for ease of identification.}
\label{tab:synonym_swap_tamil}
\end{table*}

\end{document}